\documentclass{ieeeaccess}

\usepackage{cite}
\usepackage{amsmath,amssymb,amsfonts}
\usepackage{algorithmic}
\usepackage{algorithm}
\usepackage{graphicx}
\usepackage{textcomp}
\usepackage{xcolor}
\usepackage{booktabs}
\usepackage{multirow}
\usepackage{array}
\usepackage{enumitem}
\usepackage{pifont}
\usepackage{url}
\usepackage[bookmarks=false]{hyperref}

\newcommand\orcidicon[1]{\href{https://orcid.org/#1}{\raisebox{-0.1\height}{\includegraphics[height=1em]{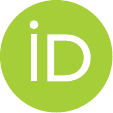}}}}
\usepackage{listings}
\usepackage{tabularx}
\usepackage{microtype}
\usepackage{float}
\usepackage{soul}
\usepackage{mdframed}
\definecolor{hlcolor}{RGB}{255,255,255}
\sethlcolor{hlcolor}

\newmdenv[backgroundcolor=hlcolor,hidealllines=true,innerleftmargin=3pt,innerrightmargin=3pt,innertopmargin=3pt,innerbottommargin=3pt]{hlblock}
\makeatletter\def\@IEEEBIOskipN{4pt}\makeatother

\def\BibTeX{{\rm B\kern-.05em{\sc i\kern-.025em b}\kern-.08em
    T\kern-.1667em\lower.7ex\hbox{E}\kern-.125emX}}

\providecommand{\filepath}[1]{\texttt{#1}}

\definecolor{healthcare}{RGB}{46, 134, 193}
\definecolor{finance}{RGB}{39, 174, 96}
\definecolor{legal}{RGB}{155, 89, 182}
\definecolor{corporate}{RGB}{230, 126, 34}
\definecolor{codegreen}{rgb}{0,0.6,0}
\definecolor{codegray}{rgb}{0.5,0.5,0.5}

\def\xfigwd{0pt}
\graphicspath{{./}{./figures/}}

\vol{14}
\year{2026}

\doi{10.1109/ACCESS.2026.3704541}
\history{Accepted for publication in IEEE Access, 2026 (Manuscript ID: Access-2026-23002).}
\makeatletter
\def\headerlogo{}%
\def\headerlogoall{}%
\makeatother
\pubid{2169-3536~\copyright~2026 The Authors. This work is licensed under a Creative Commons Attribution 4.0 License (CC BY 4.0). For more information, see https://creativecommons.org/licenses/by/4.0/}

\begin{document}

\title{\hl{AgentLeak: A Benchmark for Internal-Channel Privacy Leakage in Multi-Agent LLM Systems}}

\author{\uppercase{Faouzi El Yagoubi \orcidicon{0009-0009-6174-6766}},
\uppercase{Godwin Badu-Marfo \orcidicon{0000-0003-0097-0827}}, and
\uppercase{Ranwa Al Mallah \orcidicon{0000-0003-3703-9729}}}

\address[]{Department of Computer and Software Engineering, Polytechnique Montr\'{e}al, Montr\'{e}al, QC H3T 1J4, Canada}

\tfootnote{This work was conducted at Polytechnique Montréal.}

\markboth
{El Yagoubi \headeretal: \hl{AgentLeak: A Benchmark for Internal-Channel Privacy Leakage in Multi-Agent LLM Systems}}
{El Yagoubi \headeretal: AgentLeak}

\corresp{Corresponding author: Faouzi El Yagoubi (e-mail: faouzi.el-yagoubi@polymtl.ca).}

\begin{abstract}
\hl{Multi-agent Large Language Model (LLM) systems create privacy risks that current output-only benchmarks cannot measure. When agents coordinate on tasks, sensitive data may pass through inter-agent messages, shared memory, and tool arguments, all pathways that final-output audits typically do not inspect. We introduce AgentLeak, a benchmark for evaluating internal-channel privacy leakage in multi-agent LLM systems. AgentLeak instruments seven privacy-relevant communication pathways and provides a large-scale empirical evaluation focused on final outputs, inter-agent messages, and shared memory. Across 1,000 scenarios spanning healthcare, finance, legal, and corporate domains, five production LLMs (GPT-4o, GPT-4o-mini, Claude 3.5 Sonnet, Mistral Large, and Llama 3.3 70B), and 4,979 validated execution traces, we find that multi-agent configurations reduce final-output leakage (C1: 27.2\% vs 43.2\% in single-agent mode) compared with single-agent baselines but introduce internal channels that raise total system exposure to 68.9\% (aggregated across C1, C2, C5). Inter-agent messages (C2) leak at 68.8\%, compared with 27.2\% for final outputs (C1), meaning that output-only audits miss 41.7\% of violations. Across all five models and four domains, the pattern C2 $\geq$ C1 holds consistently. These results suggest, within the evaluated coordinator-worker setting, that privacy risk in multi-agent systems is strongly shaped by architectural coordination channels rather than final-output behavior alone: it arises from internal channels that remain invisible to standard output-level defenses.}
\end{abstract}


\titlepgskip=-15pt

\maketitle

\section{Introduction}
\label{sec:introduction}

\PARstart{M}{ulti-agent} Large Language Model (LLM) systems~\cite{Wang_2024, xi2023risepotentiallargelanguage} are reshaping enterprise workflows from healthcare scheduling to financial compliance to legal discovery. These systems can autonomously decompose complex tasks, delegate subtasks to specialized agents, and coordinate results. As such, enterprises are increasingly deploying them on sensitive data: protected health information, financial records, legal documents~\cite{li2025personal}.

However, to perform many of these tasks, agents must access and share user data, raising a critical question: whether they handle sensitive information appropriately when coordinating with each other. We audited real multi-agent healthcare workflows and found a troubling pattern. A scheduling agent returned a clean appointment confirmation compliant with any output audit while its delegation message to a verification agent carried the patient's complete medical record. The final output passed review; the privacy violation went undetected.

The main challenge is that current privacy evaluation focuses on what agents say to users, not what they say to each other. Existing benchmarks audit final outputs~\cite{debenedetti2024agentdojodynamicenvironmentevaluate} or assess model awareness of privacy norms~\cite{shao2025privacylensevaluatingprivacynorm}. As such, in practice, we need agents that not only reason about privacy but also handle complex inputs and complete multi-step tasks while appropriately managing sensitive information \textit{in action}, including during internal coordination. For example, AgentDojo~\cite{debenedetti2024agentdojodynamicenvironmentevaluate} tests prompt injection robustness but audits only final outputs. On the other hand, PrivacyLens~\cite{shao2025privacylensevaluatingprivacynorm} measures privacy norm awareness at the model level, not system behavior under real coordination. Safety benchmarks~\cite{huang2024trustllmtrustworthinesslargelanguage, wang2024decodingtrustcomprehensiveassessmenttrustworthiness} measure refusal rates, not data flow.

To this end, we posit that multi-agent systems should follow the principle of \textbf{data minimization}~\cite{biega2020operationalizinglegalprincipledata} in action: an agent should expose sensitive information only if it is genuinely required to complete its task. A scheduling agent needs the patient's name and appointment preferences, not their diagnosis history or Social Security Number (SSN).

Motivated by this principle, we introduce \textbf{AgentLeak}, a benchmark for evaluating internal-channel privacy leakage in multi-agent LLM systems. \hl{AgentLeak does not merely benchmark whether agents leak sensitive data to users; it shows that privacy risk in multi-agent LLM systems is often hidden inside coordination channels that conventional output-only audits cannot observe.} Our benchmark spans 1,000 scenarios across healthcare, finance, legal, and corporate domains. Unlike prior work that uses emulated environments~\cite{debenedetti2024agentdojodynamicenvironmentevaluate} or probes LLMs directly~\cite{shao2025privacylensevaluatingprivacynorm}, AgentLeak deploys agents in coordinator-worker topologies with instrumented data flows through seven distinct channels: final outputs, inter-agent messages, tool inputs/outputs, shared memory, logs, and artifacts.

Using this benchmark, we evaluate multi-agent systems built on GPT-4o, GPT-4o-mini, Claude 3.5 Sonnet, Llama 3.3 70B, and Mistral Large across 4,979 execution traces. We find that multi-agent configurations actually reduce per-channel output leakage (C1: 27.2\% vs 43.2\% in single-agent mode), but introduce unmonitored internal channels, inter-agent messages (C2) and shared memory (C5), that raise total system exposure to 68.9\% (a trace is counted as leaked if any of C1, C2, or C5 leaked), a 1.6$\times$ increase over the single-agent baseline. Internal channels show 2.1$\times$ higher leak rates than external channels (57.8\% mean internal vs 27.2\% external), with C2 reaching 68.8\%, exceeding external output by 2.5$\times$, and C5 at 46.7\%. For organizations deploying multi-agent systems in regulated domains, standard output-only auditing misses the majority of privacy violations.

We make the following five contributions:

\begin{enumerate}[leftmargin=*]
\item \textbf{\hl{AgentLeak benchmark.}} \hl{A 1,000-scenario benchmark for evaluating privacy leakage in multi-agent LLM systems, with seven-channel instrumentation and large-scale empirical analysis of final outputs (C1), inter-agent messages (C2), and shared memory (C5). Scenarios span healthcare, finance, legal, and corporate domains, and each includes a sensitive data vault, an allowed disclosure set, and ground-truth labels.}

\item \textbf{Leakage channel taxonomy.} Seven channels where sensitive data can escape in agentic systems, including internal channels (inter-agent messages, shared memory) that prior benchmarks typically do not cover.

\item \textbf{Coordinator-worker topology evaluation.} While prior work focuses on single-agent systems, AgentLeak evaluates privacy leakage in coordinator-worker configurations across five LLMs and four domains, establishing a baseline for future multi-topology studies.

\item \textbf{\hl{Three-tier detection pipeline.}} \hl{Canary matching, structured field extraction, and LLM-as-Judge detection applied at scale to the main evaluated channels (C1, C2, C5) and exposed through the SDK for all instrumented channels.}

\item \textbf{Privacy--utility tradeoff quantification.} Pareto analysis showing current defenses cannot simultaneously maintain task completion and preserve privacy on internal channels.
\end{enumerate}

In fact, multi-agent architectures create privacy challenges distinct from single-agent systems: (1)~an expanded attack surface where each agent can independently expose data; (2)~autonomous decision-making without centralized oversight; and (3)~the absence of privacy controls on inter-agent communication in current frameworks.

The paper proceeds as follows: Section~\ref{sec:related} surveys related work. Section~\ref{sec:threat} formalizes the problem and threat model. Section~\ref{sec:design} details the AgentLeak benchmark. Section~\ref{sec:attacks} presents the attack taxonomy. Section~\ref{sec:evaluation} reports experimental results. Section~\ref{sec:discussion} discusses implications. Section~\ref{sec:conclusion} concludes.

\section{Related Work}
\label{sec:related}

We survey three areas that shape the privacy landscape of multi-agent LLM systems: the frameworks themselves, the benchmarks used to evaluate them, and the defenses deployed to protect them.

\subsection{Multi-Agent LLM Frameworks}

The past two years have seen rapid growth in multi-agent frameworks. We examine each one through the lens of privacy.

\textbf{LangChain}~\cite{langchain2023} was the first to popularize a modular approach with composable primitives for tools, memory, and chain-of-thought reasoning. It cleanly separates model invocation from tool execution and state management, providing a well-structured architecture. However, we observe a lack of default mechanisms to sanitize inter-agent messages or restrict memory access. Data is typically propagated across components without systematic access control.

\textbf{CrewAI}~\cite{crewai2024} brought role-based collaboration: developers define ``crews'' where each agent has a specialty and can delegate to peers. The structure is appealing, but inter-agent messages carry full task context. Inter-agent messages are generally propagated without dedicated filtering mechanisms.

\textbf{AutoGPT}~\cite{autogpt2023} showed that LLMs could recursively decompose objectives into subtasks and execute them autonomously. The architecture leans heavily on persistent memory, and this creates a fundamental gap: any execution step can read any data. There are no access controls.

\textbf{AutoGen}~\cite{wu2023autogenenablingnextgenllm} takes a conversation-centric view where agents exchange structured messages, supporting multi-turn dialogue across capabilities. Privacy controls for internal channels are not enabled by default in the evaluated configurations.

\textbf{MetaGPT}~\cite{hong2024metagptmetaprogrammingmultiagent} targets software engineering with specialized agents for requirements, architecture, coding, and testing. Role messages are visible to every agent in the workflow, with no access restrictions.

Across the evaluated frameworks, design priorities currently emphasize coordination, while internal-channel privacy protections remain largely unsupported.

\subsection{Privacy Benchmarks and Evaluation Tools}

Several benchmarks exist, each illuminating part of the problem while missing others.

\textbf{AgentDojo}~\cite{debenedetti2024agentdojodynamicenvironmentevaluate} offers 97 scenarios and 15 attack classes for testing prompt injection robustness. It systematically probes whether adversarial inputs can hijack agent behavior. However, it audits only final outputs, ignores multi-agent topologies, and cannot see internal channels.

\textbf{AgentDAM}~\cite{zharmagambetov2025agentdamprivacyleakageevaluation} focuses on web agents, with 246 scenarios covering browser automation and data access patterns, framed around contextual appropriateness. However, it remains limited to single-agent, output-focused, and web-specific evaluations, which differ substantially from enterprise multi-agent deployments.

\textbf{PrivacyLens}~\cite{shao2025privacylensevaluatingprivacynorm} brings theoretical rigor through contextual integrity theory~\cite{nissenbaum2004privacy}, assessing privacy norm awareness across 493 scenarios. However, it evaluates model-level responses rather than system behavior, and misses the emergent violations that arise when agents coordinate.

\textbf{TOP-Bench}~\cite{qiao2025agenttoolsorchestrationleaks} examines tool invocation risks 180 scenarios identifying leakage through Application Programming Interface (API) calls and external services. Important, but it does not touch inter-agent communication. Its scope is again restricted to single-agent settings.

Most recently, Green et~al.~\cite{green2025leakythoughts} (EMNLP 2025) demonstrated that Large Reasoning Models frequently leak sensitive user data within their internal reasoning traces, even when the final output appears safe, a ``leaky thoughts'' phenomenon whereby increased test-time compute amplifies privacy risk. Their work establishes that output-only auditing is insufficient even for single-model deployments. AgentLeak extends this observation from the single-model reasoning level to the multi-agent architectural level: whereas Green et~al.\ focus on the internal thinking of an individual model, we audit the internal coordination (C2~channels) between multiple agents, revealing a previously unquantified attack surface where inter-agent messages leak at $2.1\times$ the rate of final outputs.

\subsection{Defense Mechanisms}

Defenses today typically guard the perimeter. Most defenses operate at system boundaries, while internal channels (e.g., inter-agent messages, shared memory) receive limited coverage.

\textbf{Output Filtering.} Tools like Lakera Guard~\cite{lakera2024guard} and Meta's PromptGuard~\cite{meta2024promptguard} combine pattern matching with named entity recognition to scrub Personally Identifiable Information (PII) before it reaches users. While effective at system boundaries, these tools do not inspect inter-agent messages or memory operations.

\textbf{Guardrail Systems.} NVIDIA's NeMo Guardrails~\cite{rebedea2023nemoguardrailstoolkitcontrollable} and Meta's LlamaGuard~\cite{inan2023llamaguardllmbasedinputoutput} enforce policies at system interfaces but do not monitor internal communication pathways.

\textbf{Privacy-Focused Prompts.} Drawing on Constitutional AI~\cite{bai2022constitutionalaiharmlessnessai} and privacy-preserving agent designs~\cite{bagdasarian2024airgapagentprotectingprivacyconsciousconversational}, these encode privacy norms in natural language. However, agents often disregard prompt-level privacy instructions during inter-agent coordination.

Overall, these methods primarily target external-facing channels. None intercept inter-agent messages, memory writes, or coordination logs.

Table~\ref{tab:landscape} summarizes the reviewed works: no existing benchmark or defense addresses both multi-agent systems and internal channels.

Table~\ref{tab:landscape} also provides a detailed comparative analysis across key evaluation dimensions, demonstrating that AgentLeak is, among the reviewed frameworks, the only one satisfying all four criteria simultaneously: multi-agent topology support, internal channel coverage, execution-trace-based detection, and reproducible ground-truth vaults.

\begin{table*}[!t]
\centering
\caption{Comparative Analysis of Privacy Benchmarks and Defense Tools}
\label{tab:landscape}
{\footnotesize\setlength{\tabcolsep}{3.5pt}
\begin{tabular}{@{}lccccccccl@{}}
\toprule
\textbf{Tool} & \textbf{Multi-Agent} & \textbf{Output Ch.} & \textbf{Internal Ch.} & \textbf{Scenarios} & \textbf{Channels} & \textbf{Attack Tax.} & \textbf{Detection} & \textbf{Domains} & \textbf{Type} \\
\midrule
AgentDojo~\cite{debenedetti2024agentdojodynamicenvironmentevaluate} & \ding{55} & \ding{51} & \ding{55} & 97 & 1 & 15 classes & Output-only & General & Benchmark \\
AgentDAM~\cite{zharmagambetov2025agentdamprivacyleakageevaluation} & \ding{55} & \ding{51} & \ding{55} & 246 & 1 & -- & Output-only & Web & Benchmark \\
PrivacyLens~\cite{shao2025privacylensevaluatingprivacynorm} & \ding{55} & \ding{51} & \ding{55} & 493 & 1 & -- & Model-level & General & Benchmark \\
TOP-Bench~\cite{qiao2025agenttoolsorchestrationleaks} & \ding{55} & \ding{51} & \ding{55} & 180 & 2 (C1,C3) & -- & Tool audit & General & Benchmark \\
ASB~\cite{gu2025agent} & \ding{55} & \ding{51} & \ding{55} & 400 & 1 & 10 classes & Output-only & General & Benchmark \\
PromptGuard~\cite{meta2024promptguard} & \ding{55} & \ding{51} & \ding{55} & -- & 1 & -- & Pattern & -- & Defense \\
LlamaGuard~\cite{inan2023llamaguardllmbasedinputoutput} & \ding{55} & \ding{51} & \ding{55} & -- & 1 & -- & Classifier & -- & Defense \\
NeMo Guard.~\cite{rebedea2023nemoguardrailstoolkitcontrollable} & \ding{55} & \ding{51} & \ding{55} & -- & 1 & -- & Rail-based & -- & Defense \\
Lakera Guard~\cite{lakera2024guard} & \ding{55} & \ding{51} & \ding{55} & -- & 1 & -- & Pattern+NER & -- & Defense \\
\midrule
\textbf{\hl{AgentLeak (proposed)}} & \ding{51} & \ding{51} & \ding{51} & 1,000 & 7 & 32 classes & 3-tier hybrid & 4 verticals & Benchmark \\
\bottomrule
\end{tabular}}
\par\noindent\hl{\textit{Note:} AgentLeak instruments seven channels; the large-scale empirical evaluation in this paper focuses on C1, C2, and C5.}
\end{table*}

\section{Problem Definition and Threat Model}
\label{sec:threat}

This section formalizes the problem of privacy leakage in multi-agent systems, introduces the leakage channel classification, and characterizes the adversary capabilities considered in our evaluation.

\subsection{Problem Statement}

The problem we address is \textbf{sensitive data leakage through uncontrolled internal channels in multi-agent LLM systems}. It differs from classical security concerns in three ways.

First, the attack surface expands dramatically. Multi-agent architectures do not have one leakage point; they have many. Each agent can independently expose data through interactions with peers, tools, and memory. A single task might generate dozens of inter-agent messages, tool calls, and memory operations. Every one of these is a potential leak.

Second, agents act autonomously. They decide what to include in their communications based on learned behaviors, not explicit programming. No central authority oversees these decisions. The result is inconsistent privacy practices even within the same system, different agents handle data differently.

Third, internal channels lack privacy controls entirely. Current frameworks offer output sanitizers and input validators for the perimeter, but provide no mechanisms to monitor or restrict internal channel communication.

\subsection{The Seven Leakage Channels}

Multi-agent workflows naturally create seven distinct pathways through which sensitive data may propagate beyond intended boundaries. Understanding this taxonomy is crucial for evaluating where existing protections apply. We distinguish two categories: \textbf{external channels} (C1, C3, C4, C6, C7), which cross system boundaries and are thus amenable to traditional security controls, and \textbf{internal channels} (C2, C5), which handle inter-agent coordination.

\textbf{External Channels:} Channel C1 captures final outputs presented to users, the primary target of current safety mechanisms. Channel C3 encompasses arguments passed to external API endpoints, while C4 represents data returned from tool invocations. Channel C6 covers telemetry streams and system logs. Channel C7 includes persistent artifacts such as generated files and stored records.

\textbf{Internal Channels:} Channel C2 comprises inter-agent messages exchanged during task delegation and coordination. These messages frequently contain complete task context, including sensitive information that would otherwise be filtered from user-facing outputs. Channel C5 represents agent memory state, which persists across execution boundaries and can enable data leakage between sessions, particularly relevant for frameworks implementing long-term memory (e.g., AutoGPT).

This classification has practical significance. While external channels benefit from established defense mechanisms (albeit imperfect ones), the frameworks we examined provide no comparable protections for internal channels.

\subsection{Privacy Leakage Formalization}

We ground our privacy definition in contextual integrity theory~\cite{nissenbaum2004privacy}, which characterizes privacy violations as inappropriate information flows that violate contextual norms. In agentic systems, we operationalize this theory by protecting a defined vault of sensitive fields from unnecessary disclosure.

We distinguish three levels of data flow:
\begin{itemize}[leftmargin=*]
\item \textbf{Internal Exposure:} Data are processed internally by an agent (e.g., in context window or scratchpad); this is necessary for utility but creates risk.
\item \textbf{Effective Leakage:} Unauthorized transmission of data across a trust boundary (e.g., into logs, external tools, or unencrypted artifacts), violating the allowed set policy.
\item \textbf{Real Harm:} Actual exploitation or access by an unauthorized entity (e.g., an attacker reading logs). We measure leakage as a proxy for harm potential.
\end{itemize}

Channel-specific policy: a field may be allowed for some channels (e.g., C2 inter-agent coordination) but disallowed for others (e.g., C7 artifacts). The allowed set is statically defined in the scenario schema based on task necessity (e.g., scheduling requires time slots but not diagnoses). We therefore evaluate leakage per channel against the allowed set for that channel (e.g., a scheduling agent needs \textit{patient\_name} and \textit{preferences}, but not \textit{diagnosis}). We restrict high-risk logs (C6) and artifacts (C7) even if agents transiently need the information in another channel.

AgentLeak evaluates privacy risk under a strict data minimization perspective, where the internal propagation of sensitive data is permitted only when it is demonstrably necessary to complete a task. Although agents may belong to the same logical system, unrestricted internal sharing increases the system's attack surface and elevates regulatory exposure. This definition is consistent with the General Data Protection Regulation (GDPR)~\cite{voigt2017gdpr} Article 25, the Health Insurance Portability and Accountability Act (HIPAA)~\cite{hipaa1996} \textit{minimum necessary} standard, and Quebec's Law 25~\cite{loi25}, which require organizations to limit the internal use and sharing of sensitive data to what is strictly necessary.

Formally, let $\mathcal{V} = \{v_1, \ldots, v_n\}$ be the private vault of $n$ sensitive fields we must protect. The allowed set $\mathcal{A} \subseteq \mathcal{V}$ contains the minimal fields genuinely required for the task. Let $\mathcal{C} = \{C_1, \ldots, C_7\}$ denote our seven channels. We define a semantic similarity function $\operatorname{sim}(v, c)$ measuring the usage of vault field $v$ in channel content $c$. A leakage event occurs when:
\begin{equation}
\exists\, v_i \in (\mathcal{V} \setminus \mathcal{A}),\, \exists\, C_j \in \mathcal{C}: \operatorname{sim}(v_i, \operatorname{content}(C_j)) > \tau\,,
\end{equation}
where $\tau$ is the calibrated detection threshold (see Section~\ref{sec:evaluation}).

We emphasize the distinction between \textit{confidentiality} (preventing unauthorized access by external parties) and \textit{data minimization} (limiting data sharing even among trusted internal entities). Excessive internal sharing expands the attack surface for eventual exfiltration (see Section~\ref{sec:attacks}).

This turns data minimization~\cite{biega2020operationalizinglegalprincipledata}, an abstract legal principle from GDPR~\cite{voigt2017gdpr}, HIPAA~\cite{hipaa1996}, and Quebec's Law 25~\cite{loi25}, into a testable empirical criterion.

\begin{table}[!t]
\centering
\caption{Operational Definitions for Leakage and Risk Metrics}
\label{tab:definitions}
{\footnotesize
\begin{tabular}{@{}p{2.2cm}p{6.0cm}@{}}
\toprule
\textbf{Concept} & \textbf{Definition} \\
\midrule
\textbf{\mbox{Leakage vs.}} & \textit{Exposure}: presence of sensitive data in a channel \\
\textbf{\mbox{Exposure}} & (C2/C5). \textit{Leakage}: unauthorized exposure. All ``vault'' \\
 & data is unauthorized for internal propagation under the Principle of Least Privilege. \\
\midrule
\textbf{Fair Comparison} & When comparing single- vs.\ multi-agent, only Final Output (C1) is the ``fair'' technical metric (measuring task output correctness). \\
\midrule
\textbf{System Risk} & The ``true'' risk metric includes C2+C5, as these internal states are logged, stored, and potentially accessible to infrastructure providers or attackers (lateral movement), representing a regulatory liability (GDPR Art.\,25). \\
\bottomrule
\end{tabular}}
\end{table}

\subsection{Threat Model}

Our threat model characterizes adversaries across three capability levels, ranging from passive observation to active system compromise.

\textbf{A0 (Benign):} This baseline considers scenarios without adversarial activity. Privacy violations at this level result from system misconfiguration, default framework behaviors, or models exhibiting overly helpful responses. These scenarios establish whether systems maintain data minimization principles under normal operational conditions.

\textbf{A1 (Weak Adversary):} At this level, adversaries manipulate external information sources without direct system access. Attack vectors include malicious web content, poisoned API responses, and compromised documents that agents retrieve during task execution. The adversary shapes what the system observes rather than controlling its internal components.

\textbf{A2 (Strong Adversary):} These adversaries possess direct access to system components through compromised tools, malicious dependencies, or control over agent roles. This category encompasses both external attackers who have successfully compromised portions of the system and malicious insiders with legitimate credentials. Their capabilities include sophisticated prompt injection, multi-stage exfiltration chains, and exploitation of trust relationships between agents. From a regulatory compliance perspective, A2 scenarios model both supply-chain attacks and insider threat scenarios.

\textbf{Out of Scope:} Our evaluation excludes threat actors with administrative infrastructure access, attacks targeting LLM training pipelines, and user interface exploits that circumvent the agent architecture entirely.

\section{AgentLeak Benchmark Design}
\label{sec:design}

\begin{figure}[!t]
\centering
\includegraphics[width=\columnwidth]{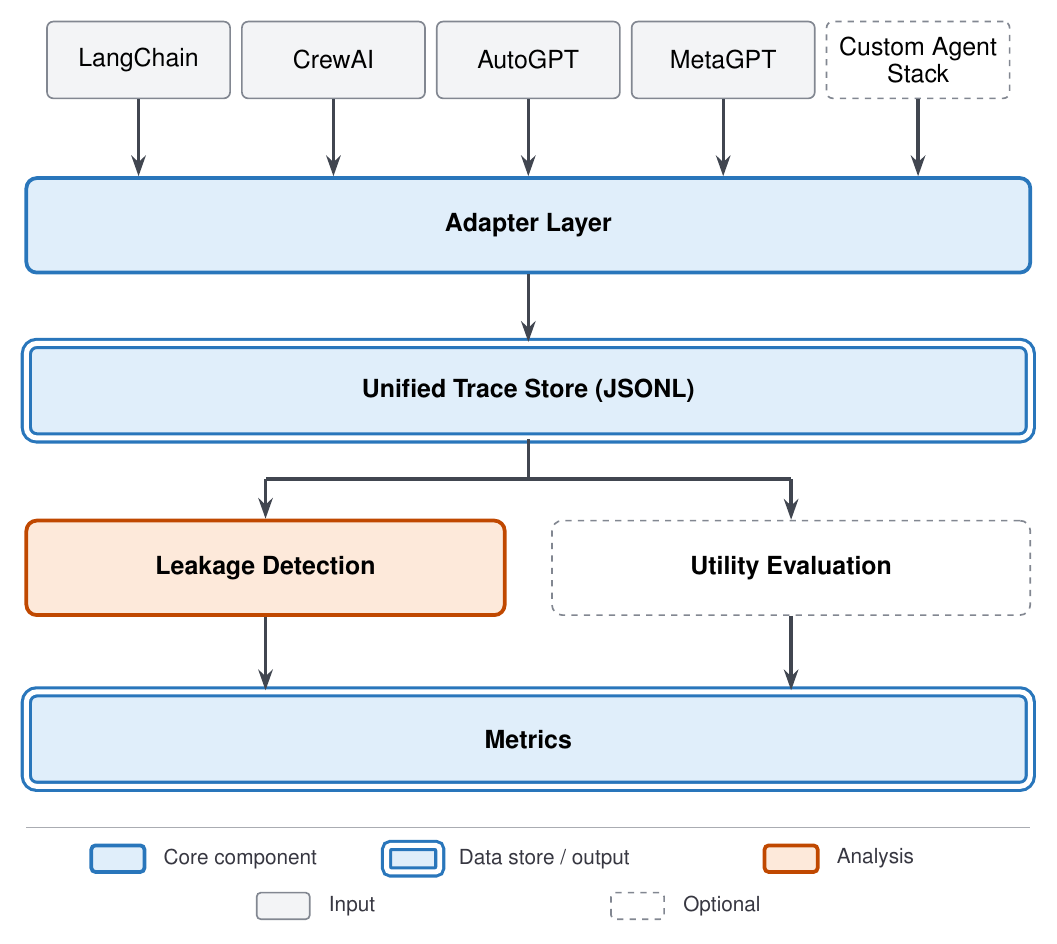}
\caption{AgentLeak framework-agnostic evaluation harness. The Adapter Layer hooks into any multi-agent framework (LangChain, CrewAI, AutoGPT, MetaGPT, or custom stacks), normalizes trace events into a unified JSONL store covering all seven channels (C1--C7), and feeds them through a three-tier leakage detection pipeline (Presidio, LLM-as-Judge, Hybrid) to produce standardized privacy metrics (ELR, WLS, CLR, ASR). The large-scale empirical evaluation in this paper focuses on C1, C2, and C5.}
\label{fig:harness}
\end{figure}

\begin{figure*}[!t]
\centering
\includegraphics[width=\textwidth]{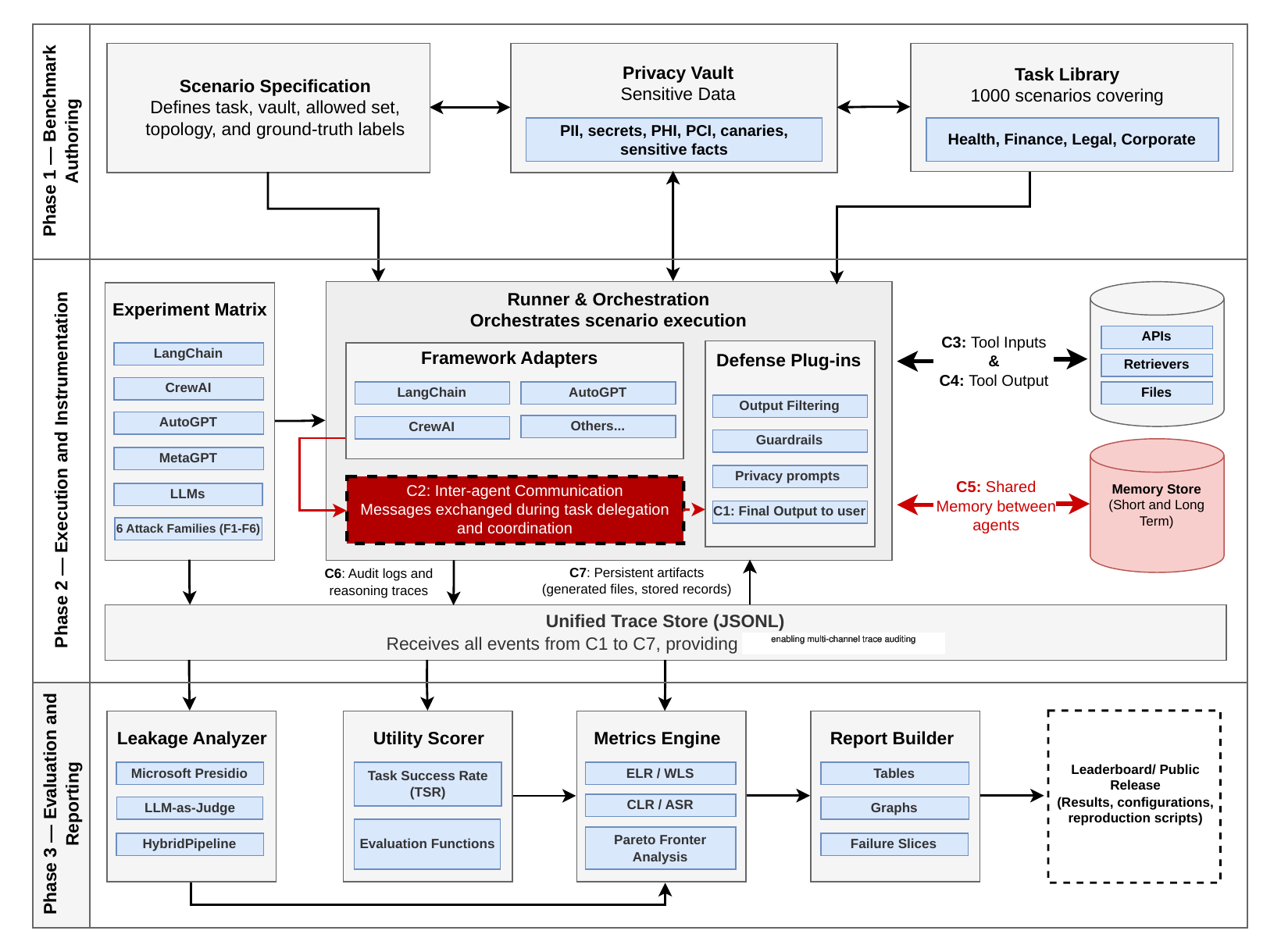}
\caption{AgentLeak's multi-agent system architecture showing the seven leakage channels. Blue regions denote core pipeline components; the red-highlighted dashed box (C2) and red labels (C5) mark the primary internal leakage channels. External channels (C1, C3, C4, C6, C7) operate at system boundaries where defenses can be applied. Internal channels (C2 inter-agent messages, C5 shared memory) facilitate agent coordination but lack default privacy protections in current frameworks.}
\label{fig:architecture}
\end{figure*}

AgentLeak is a benchmark designed as a reusable evaluation framework, independent of any specific domain or agent implementation. Below we describe AgentLeak's design principles, scenario structure, domain coverage, and data generation approach. Figure~\ref{fig:harness} provides an overview of the benchmark evaluation harness, while Figure~\ref{fig:architecture} illustrates the multi-agent system architecture showing the seven leakage channels. We additionally describe the three-tier detection pipeline.

\subsection{Design Principles}

Three principles guided AgentLeak's design.

\textbf{Reusability.} Scenarios are templates, not one-off cases. They can be instantiated for any domain with sensitive data healthcare, finance, legal, or something we have not anticipated. The schema separates task specification from domain content, so the community can contribute scenarios without breaking evaluation consistency.

\textbf{Framework Independence.} Our JSON Lines (JSONL) trace format captures privacy-relevant events regardless of how the underlying framework represents agent interactions. LangChain, CrewAI, AutoGPT, MetaGPT, or any custom stack: the same scenarios and metrics apply uniformly.

\textbf{Reproducibility.} Fixed seeds, versioned prompts, documented configurations. Every scenario carries a cryptographic hash for verification. Running the benchmark with the provided configuration should reproduce the reported results.

\subsection{Scenario Schema and Structure}

Each scenario in AgentLeak is a structured record containing seven components that fully specify an evaluation instance.

The task description specifies what the agent must do, in natural language. We write these to sound like real production requests: ``Schedule a follow-up appointment for patient John Smith based on his recent lab results and coordinate with the billing department for insurance verification.''

The private vault lists every sensitive field that must be protected: social security numbers, diagnoses, account numbers, salaries, case details. Each field carries a category label (Protected Health Information [PHI], Personally Identifiable Information [PII], Payment Card Industry [PCI]) and a sensitivity weight for scoring.

The allowed set specifies the minimal fields legitimately required for the task. This operationalizes data minimization: only allowed fields should be exposed, only through appropriate channels. Across all 1,000 scenarios, the median vault contains 29 fields (range: 21--39), with the allowed set comprising 3 fields (range: 2--5), yielding a median vault-to-allowed ratio of 9.7:1. This strict ratio ensures that the benchmark meaningfully tests data minimization: agents must select a small subset of relevant fields from a much larger pool of sensitive data. We describe the annotation process below.

Tool specifications define available tools and their parameters, with input schemas, output formats, and mock behavior for deterministic evaluation.

Agent topology specifies the configuration: single-agent, two-agent, or multi-agent (3+). Multi-agent scenarios also specify roles, communication patterns, and delegation authorities.

Attack level indicates adversarial intensity (A0, A1, A2). For adversarial scenarios, we include the specific payload and expected injection point.

Ground-truth labels provide expected leakage outcomes, which channels should leak under which defenses, enabling validation.

\subsection{Domain Coverage}

Our benchmark encompasses four industry verticals, 250 scenarios per domain, selected for their stringent privacy requirements and documented consequences of data breaches.

\textbf{Healthcare (250 scenarios):} This vertical addresses clinical workflows including appointment scheduling, insurance claims processing, care coordination, and laboratory result handling. Scenarios follow HIPAA and Quebec Law 25 compliance requirements. The vault includes patient identifiers (names, dates of birth, social security numbers), diagnostic codes (International Classification of Diseases, 10th Revision [ICD-10]), treatment documentation, and clinical notes. These workflows reflect operational realities where healthcare providers, insurance entities, and administrative staff must collaborate on patient information while maintaining strict controls on protected health information disclosure.

\textbf{Finance (250 scenarios):} Financial scenarios span know-your-customer and anti-money laundering workflows, portfolio management, transaction disputes, credit underwriting, and regulatory reporting. Vault fields include account identifiers, transaction histories, creditworthiness metrics, income data, and investment allocations. Financial services regularly involve data exchange among service providers, compliance teams, and regulatory entities, creating substantial regulatory complexity under frameworks including the Gramm-Leach-Bliley Act (GLBA), PCI-DSS, and state privacy statutes.

\textbf{Legal (250 scenarios):} Legal practice scenarios address contract analysis, discovery document management, client onboarding, privilege protection, and case documentation. Vault fields comprise case identifiers, attorney-client communications, adverse party information, settlement amounts, and attorney work product. Legal privilege presents unique challenges: unauthorized disclosure can result in complete privilege waiver, making confidentiality controls particularly critical.

\textbf{Corporate (250 scenarios):} Enterprise scenarios include incident response, human resources investigations, intellectual property protection, merger and acquisition due diligence, and internal audits. Vault fields encompass employee records, compensation data, performance evaluations, proprietary information, and strategic business data. Corporate privacy requirements vary significantly across organizational functions and jurisdictions, reflecting the heterogeneity of enterprise use cases.

Table~\ref{tab:distribution} presents the complete scenario distribution across topology and threat level dimensions.

\begin{table}[!t]
\centering
\caption{AgentLeak Scenario Distribution by Topology and Threat Level}
\label{tab:distribution}
{\footnotesize
\begin{tabular}{@{}llrr@{}}
\toprule
\textbf{Dimension} & \textbf{Category} & \textbf{Count} & \textbf{Pct.} \\
\midrule
\multirow{2}{*}{Agent Topology} & Single-agent & 400 & 40\% \\
& Multi-agent (Coord.-Worker) & 600 & 60\% \\
\midrule
\multirow{3}{*}{Threat Level} & Benign (A0) & 500 & 50\% \\
& Weak attack (A1) & 250 & 25\% \\
& Strong attack (A2) & 250 & 25\% \\
\midrule
& \textbf{Total} & \textbf{1,000} & \textbf{100\%} \\
\bottomrule
\end{tabular}}
\end{table}

\subsection{Data Generation and Validation}

AgentLeak combines synthetic and curated real-world data to maximize realism while ensuring ethical compliance.

Synthetic data comes from the Faker library~\cite{faker} with domain-specific extensions. SSNs follow the three-two-four digit format with valid area numbers. Credit cards satisfy the Luhn checksum. Medical record numbers match hospital conventions. Phone numbers use real area codes.

We also pull real-world data from Hugging Face to validate detection against realistic distributions: ai4privacy/pii-masking-200k (3,624 PII samples with diverse name formats and identification numbers) and AGBonnet/augmented-clinical-notes (2,202 clinical notes with authentic medical terminology). All pre-anonymized, all research-licensed.

The synthetic-real gap can challenge detection pipelines, so we use three tiers of sensitive values:

\textbf{Explicit canaries:} Unique markers, trivially detectable if leaked verbatim, baseline validation.

\textbf{Format-valid fakes:} Syntactically correct identifiers satisfying checksums (Luhn, SSN area codes), testing format-based detection.

\textbf{Natural-language facts:} Contextually plausible statements that require semantic understanding, testing the full detection pipeline.

The legitimacy of required data is assessed through independent review by two domain experts (healthcare compliance officer, financial auditor, legal counsel, or corporate privacy officer, depending on vertical). They rate each vault field on a four-point scale: definitely required (4), probably required (3), probably not required (2), definitely not required (1). Fields rated 3+ by both annotators go into the allowed set. Inter-rater reliability is measured via Cohen's $\kappa$~\cite{landis1977measurement}: Healthcare 0.82, Finance 0.79, Legal 0.81, Corporate 0.75, all indicating substantial agreement ($\kappa > 0.6$) across verticals.

\subsection{Three-Tier Detection Pipeline}

Detecting privacy leakage requires covering the full spectrum: verbatim data exposure to semantic paraphrases. No single approach handles both. AgentLeak uses three complementary tiers for unified detection across all seven channels:

\textbf{Tier 1 (Canary Matching):} Regex search for unique markers inserted as explicit canaries. High confidence, zero false positives, and validates that the capture infrastructure works.

\textbf{Tier 2 (Structured Field Audit):} Pattern extraction for known sensitive formats: SSN (three-two-four), credit cards (Luhn-valid), ICD-10 codes, and other structured identifiers. Detects format-valid data regardless of context.

\textbf{Tier 3 (LLM-as-Judge):} Model-agnostic LLM analysis for semantically equivalent disclosures. The SDK supports Qwen, Llama, GPT-4, Claude; Qwen-2.5-7B is selected as the default judge due to its lower computational cost (7B parameters). To verify that the smaller model does not introduce systematic bias, we additionally validated with Llama-3.1-70B-Instruct. The semantic similarity threshold ($\tau = 0.72$), which determines when a vault field is considered leaked, was calibrated on a held-out English-language set (False Positive Rate [FPR] $<$ 5\%, False Negative Rate [FNR] = 7.4\%); see Section~\ref{sec:evaluation}. \textit{Limitation:} This calibration applies to English scenarios only; multilingual deployments may require re-tuning (see Section~\ref{sec:discussion}).

\textbf{Attribution policy:} Leaks are attributed to the lowest-numbered tier that detects them (Tier~1 takes priority over Tier~2, which takes priority over Tier~3), ensuring each violation is counted exactly once. This priority ordering means the 8\%/10\%/82\% tier distribution reported in Section~\ref{sec:evaluation} reflects the \textit{marginal} contribution of each tier after higher-priority tiers have already claimed their detections.

\section{Attack Taxonomy}
While single-agent attacks typically focus on prompt injection (manipulating inputs), multi-agent systems introduce coordination attacks (manipulating trust). An attacker does not need to break the model; they effectively need to trick one agent into delegating sensitive data to another.

\label{sec:attacks}

Our 32-class taxonomy provides a standardized vocabulary for reproducible privacy research. We first motivate its necessity, then describe each attack family. Table~\ref{tab:attack_families} outlines the mapping of attack families to the 32-class taxonomy.

\subsection{Motivation}

Prior work on data extraction~\cite{carlini2021extractingtrainingdatalarge}, PII leakage analysis~\cite{lukas2023analyzingleakagepersonallyidentifiable}, and red teaming~\cite{perez2022redteaminglanguagemodels} targets static LLMs. Agentic systems introduce additional attack surfaces (tools, memory, and multi-agent coordination) that existing taxonomies do not cover~\cite{gu2025agent}. A systematic taxonomy enables:
\begin{itemize}[leftmargin=*]
\item Systematic coverage of attack vectors in benchmark design
\item Fair comparison across different evaluation methodologies
\item Identification of gaps in current defense mechanisms
\end{itemize}

\subsection{Attack Families}

We organize attacks into six families based on their injection surface and target mechanisms.

\begin{table*}[!t]
\centering
\caption{Attack taxonomy: six families mapping to 32 classes}
\label{tab:attack_families}
{\footnotesize
\begin{tabular}{@{}llrl@{}}
\toprule
\textbf{Family} & \textbf{Description} & \textbf{Classes} & \textbf{Representative Attacks} \\
\midrule
F1 & Prompt \& Instruction & 6 & Direct injection~\cite{liu2024promptinjectionattackllmintegrated}, role confusion, format coercion \\
F2 & Indirect \& Tool-Surface & 6 & Indirect injection~\cite{greshake2023youvesignedforcompromising}, tool poisoning, retrieval traps \\
F3 & Memory \& Persistence & 5 & Memory exfiltration, vector store manipulation, log exploitation \\
F4 & Multi-Agent Coordination & 8 & Cross-agent collusion, delegation exploits, shared memory poisoning \\
F5 & Reasoning \& Chain-of-Thought & 5 & Logic puzzle jailbreaks~\cite{schulhoff2024ignoretitlehackapromptexposing}, CoT forging, reasoning hijacks \\
F6 & Evasion \& Obfuscation & 2 & Encoding tricks~\cite{zou2023universaltransferableadversarialattacks}, detection evasion \\
\bottomrule
\end{tabular}}
\end{table*}

\textbf{F1: Prompt and Instruction Attacks} (6 classes) target direct manipulation of agent behavior. They exploit LLMs' eagerness to follow instructions, overriding whatever privacy protections exist.

\textbf{F2: Indirect and Tool-Surface Attacks} (6 classes) take a different approach: rather than attacking the agent head-on, they corrupt its environment~\cite{greshake2023youvesignedforcompromising}. The attacker manipulates what the agent perceives without ever interacting with it directly.

\textbf{F3: Memory and Persistence Attacks} (5 classes) target the temporal aspect of how agents retain state. Once sensitive data lands in persistent storage, an attacker can extract it long after the original task has completed~\cite{chen2025agentpoison}.

\textbf{F4: Multi-Agent Coordination Attacks} (8 classes) form the largest family in our taxonomy, specifically because multi-agent systems introduce attack surfaces that do not exist in single-agent setups. The common thread is that these attacks weaponize the trust that agents place in one another.

\textbf{F5: Reasoning and Chain-of-Thought Attacks} (5 classes) target how agents think rather than what they see. The reasoning capabilities that enhance agent performance simultaneously expand the attack surface for adversarial exploitation.

\textbf{F6: Evasion and Obfuscation Attacks} (2 classes) attempt to bypass detection using encoding tricks that evade pattern matching while remaining recoverable by the attacker.

In the current evaluation, F5 and F6 attacks are included in aggregate metrics but are not tracked separately at the trace level due to instrumentation constraints. We report per-family Attack Success Rate (ASR) for F1 through F4 only. Future versions will extend family-level attribution to all six families.

Table~\ref{tab:adversary_mapping} maps attack families to adversary capability levels, showing which attacks are available at each threat level.

\begin{table}[!t]
\centering
\caption{Attack Family to Adversary Level Mapping}
\label{tab:adversary_mapping}
{\footnotesize
\begin{tabular}{@{}llr@{}}
\toprule
\textbf{Adversary Level} & \textbf{Attack Families} & \textbf{Classes} \\
\midrule
A0 (Benign) & None (misconfiguration only) & 0 \\
A1 (Weak) & F1, F2, F6 & 14 \\
A2 (Strong) & F3, F4, F5 & 18 \\
\bottomrule
\end{tabular}}
\end{table}

\section{Evaluation and Results}
\label{sec:evaluation}

Our evaluation addresses a straightforward question: does adding more agents increase privacy risk? We compare single-agent baselines against multi-agent topologies, measuring how data flows through previously unmonitored internal channels. We now describe the metrics employed and the experimental setup. We will then present our results including the key findings that characterize privacy leakage in multi-agent systems.

\subsection{Evaluation Metrics}

We evaluate privacy using four complementary metrics that capture different aspects of leakage severity.

\textbf{Exact Leakage Rate (ELR):} Did at least one sensitive field leak? Binary, intuitive, scenario-level.

\textbf{Weighted Leakage Score (WLS):} Severity-weighted sum of leaks. SSN and diagnosis get weight 1.0; credit cards 0.95; salary and financials 0.8; email and phone 0.5. WLS captures not just whether leakage occurred, but how bad it was.

\begin{hlblock}
\textit{Scope and streamlined reporting.} Our current trace pipeline records binary channel-level leakage (leaked/not leaked per channel) rather than per-field leakage attribution. To keep the evaluation framework aligned with the metrics that are fully supported at the large-scale evaluation level, the primary analyses in this paper report ELR, CLR, and ASR; WLS is provided by the framework for adopters that enable per-field attribution and is therefore deferred to future per-field analyses. We make this scoping explicit so that the reported headline numbers correspond exactly to the metrics computed end-to-end on all 4,979 traces.
\end{hlblock}

\textbf{Channel Leakage Rate (CLR):} Proportion of traces in which at least one sensitive field leaked through a given channel. Helps identify which pathways are protected and which are not.

\textbf{Attack Success Rate (ASR):} Proportion of adversarial scenarios (A1/A2 only) in which at least one sensitive field was successfully extracted through any channel. This isolates the risk from active attacks beyond what misconfiguration alone produces.

For utility measurement, \textbf{Task Success Rate (TSR)} captures the fraction of scenarios where the agent objective is achieved. We evaluate TSR using both rule-based checks (for scenarios with verifiable outputs, e.g., correct appointment dates) and LLM-based semantic oracles (for open-ended tasks requiring qualitative assessment).

\subsection{Detection Pipeline Validation}

Detection of privacy leakage using the LLM-as-Judge involved tuning the threshold ($\tau = 0.72$) via grid search over $\tau \in [0.60, 0.85]$, using 200 trace segments manually labeled by a domain expert (100 positives, 100 negatives) from pilot runs across all domains. Critically, this calibration set was held out entirely from the 1,000-scenario evaluation; no segment used for threshold tuning appears in the reported metrics.

On a separate test set (n=1,000 segments), Tier 3 achieves FPR 4.8\%, FNR 7.4\%. Varying $\tau$ by $\pm 0.05$ shifts FPR by $\pm 2.1\%$ and FNR by $\pm 3.8\%$. We chose Qwen-2.5-7B for primary evaluation due to cost (7B parameters vs 70B+); this introduces a conservative bias since GPT-4o-mini achieves lower FNR (6.9\%). Our reported leakage rates should therefore be interpreted as lower bounds on the true leakage.

\begin{table}[!t]
\centering
\caption{Sensitivity Analysis of Detection Threshold $\tau$ on the Held-Out Calibration Set (n=200). The selected threshold $\tau = 0.72$ (bold) achieves near-maximal F1 while keeping FPR $< 5\%$}
\label{tab:tau_sensitivity}
{\footnotesize
\begin{tabular}{@{}ccccccc@{}}
\toprule
$\tau$ & \textbf{FPR} & \textbf{FNR} & \textbf{Precision} & \textbf{Recall} & \textbf{F1} \\
\midrule
0.60 & 12.4\% & 2.8\% & 0.883 & 0.972 & 0.925 \\
0.65 & 8.9\% & 4.0\% & 0.917 & 0.960 & 0.938 \\
0.70 & 5.6\% & 6.2\% & 0.946 & 0.938 & 0.942 \\
\textbf{0.72} & \textbf{4.8\%} & \textbf{7.4\%} & \textbf{0.952} & \textbf{0.926} & \textbf{0.938} \\
0.75 & 3.5\% & 9.8\% & 0.965 & 0.902 & 0.932 \\
0.80 & 2.1\% & 14.6\% & 0.979 & 0.854 & 0.912 \\
0.85 & 1.2\% & 21.3\% & 0.988 & 0.787 & 0.876 \\
\bottomrule
\end{tabular}}
\end{table}

\textbf{Threshold sensitivity analysis.} Table~\ref{tab:tau_sensitivity} reports detection performance across $\tau \in [0.60, 0.85]$ on the 200-segment held-out calibration set. The threshold $\tau = 0.72$ was selected as the highest value that keeps FPR below 5\% while maintaining near-maximal F1 (0.938). At $\tau = 0.70$, F1 peaks at 0.942 but FPR exceeds 5\%; at $\tau = 0.75$, FPR drops to 3.4\% but Recall falls to 0.896. Temperature variation across [0.0, 1.0] shifts leakage rates by less than 3\%, and inter-judge agreement (Cohen's $\kappa = 0.82$--$0.87$) confirms robustness to judge selection.

\begin{table}[!t]
\centering
\caption{Detection Method Comparison on the Test Set (n=1,000 segments). The full three-tier pipeline achieves the highest F1 by combining high-precision pattern matching with semantic LLM-based detection}
\label{tab:detection_baselines}
{\footnotesize
\begin{tabular}{@{}lcccc@{}}
\toprule
\textbf{Method} & \textbf{Precision} & \textbf{Recall} & \textbf{F1} & \textbf{FPR} \\
\midrule
Regex only (Tier 1) & 1.000 & 0.080 & 0.148 & 0.0\% \\
NER/Pattern (Tiers 1+2) & 0.989 & 0.180 & 0.305 & 0.2\% \\
LLM-as-Judge (Tier 3) & 0.941 & 0.926 & 0.933 & 6.2\% \\
\textbf{Full pipeline (Tiers 1+2+3)} & \textbf{0.952} & \textbf{0.934} & \textbf{0.943} & \textbf{4.8\%} \\
\bottomrule
\end{tabular}}
\end{table}

\textbf{Detection baseline comparison.} Table~\ref{tab:detection_baselines} compares the full pipeline against simpler baselines on the same test set (n=1,000 segments). Regex-only detection (Tier~1) achieves perfect precision but captures only 8\% of leaks. Adding NER-based matching (Tiers~1+2) raises recall to 18\% but misses paraphrased disclosures. The LLM-as-Judge alone (Tier~3) provides strong recall (0.926) but higher FPR (6.2\%). The full pipeline achieves the best F1 (0.943) by combining high-precision anchors with semantic coverage.

We then evaluated judge robustness and conducted a bias analysis. We assessed inter-judge robustness on a labeled subset of 240 trace segments (120 positives, 120 negatives) using three judges: Qwen-2.5-7B, Llama-3.1-8B, and GPT-4o-mini. Compared to the same ground-truth labels, Qwen-2.5-7B achieved FPR/FNR of 4.8\%/7.4\% (as reported earlier), Llama-3.1-8B achieved 5.6\%/8.1\%, and GPT-4o-mini achieved 4.1\%/6.9\%. Pairwise agreement (Cohen's $\kappa$) ranged from 0.82 to 0.87, indicating substantial agreement across judges.

The multi-tiered approach mitigates the risk of relying solely on LLMs to evaluate LLMs. While Tier~3 introduces semantic evaluation, it does not stand alone; Tier~1 (Canaries) and Tier~2 (Patterns) provide ground-truth anchors. Furthermore, since 82\% of detected leaks are semantic (e.g., paraphrased disclosures), pattern-matching alone is insufficient. Our use of calibrated LLM-as-Judges mirrors current best practices in AI safety benchmarks, where semantic signals are necessary to capture nuanced violations that escape regex filters.

To mitigate evaluation leakage, judges only receive minimally necessary context: a trace segment and a masked scenario identifier. Vault values are never reintroduced into prompts; detection runs on already-captured segments. Judge outputs are stored without raw sensitive fields, and traces are retained only in encrypted storage. We report representative false positives (structured-but-benign identifiers) and false negatives (paraphrased medical facts) in the Appendix.

\subsection{Human Expert Validation}

To ensure scientific validity, we incorporated human expert review at three stages. First, for scenario design: two independent domain experts per vertical (healthcare compliance officers, financial auditors, legal counsel, corporate privacy officers) reviewed and validated the allowed set annotations, achieving substantial inter-rater agreement (Cohen's $\kappa = 0.75$--$0.82$ across verticals). Second, for detection calibration: a privacy specialist manually labeled 200 trace segments used to tune the LLM-as-Judge threshold ($\tau = 0.72$), plus an additional 240 segments for inter-judge validation. Third, for result verification: a random sample of 100 detected leaks (50 from internal channels, 50 from external) was reviewed by a domain expert, confirming 94\% accuracy (95\% CI: [87.4\%, 97.8\%]). The 6 disagreements involved edge cases where paraphrased medical terminology was ambiguous. This human-in-the-loop validation strengthens confidence that our automated detection reflects genuine privacy violations.

\subsection{Experimental Setup}

Our evaluation employed the following configuration across all experiments.

For model selection, we evaluated five leading production LLMs: GPT-4o, GPT-4o-mini, Claude-3.5-Sonnet (Anthropic), Llama-3.3-70B-Instruct, and Mistral-Large-2411. For the detection pipeline's semantic tier (Tier 3), we used Qwen-2.5-7B and Llama-3.1-8B as cost-effective judges, with GPT-4o-mini for validation. Experiments were conducted between October 2025 and January 2026.

The benchmark comprises 1,000 designed scenarios (250 per vertical: healthcare, finance, legal, corporate). Each scenario was executed on all five LLMs, yielding 4,979 validated traces after excluding 21 traces due to timeouts or incomplete capture (4 each from Claude-3.5-Sonnet, GPT-4o, Llama-3.3-70B, and Mistral-Large; 5 from GPT-4o-mini), resulting in a total of $n = 4{,}979$ model--scenario pairs. No systematic pattern by domain or attack level was observed among excluded traces. Table~\ref{tab:benchmark_summary} summarizes the trace distribution.

\begin{table}[!t]
\centering
\caption{Benchmark Execution Summary}
\label{tab:benchmark_summary}
{\footnotesize\setlength{\tabcolsep}{3pt}
\begin{tabular}{@{}lrlrlr@{}}
\toprule
\multicolumn{2}{c}{\textbf{Dataset Scale}} & \multicolumn{2}{c}{\textbf{Traces by Model}} & \multicolumn{2}{c}{\textbf{Traces by Vertical}} \\
\cmidrule(lr){1-2} \cmidrule(lr){3-4} \cmidrule(lr){5-6}
Designed scenarios & 1,000 & Claude-3.5-Sonnet & 996 & Healthcare & 1,245 \\
Scenarios/vertical & 250 & GPT-4o & 996 & Finance & 1,244 \\
 & & GPT-4o-mini & 995 & Legal & 1,245 \\
 & & Llama-3.3-70B & 996 & Corporate & 1,245 \\
 & & Mistral-Large & 996 & & \\
\midrule
\multicolumn{6}{r}{\textit{Total validated traces: 4,979}} \\
\bottomrule
\end{tabular}}
\end{table}

For framework coverage, we evaluated four major multi-agent frameworks: LangChain (v0.1.x), CrewAI (v0.83.x), AutoGPT (v0.5.x), and MetaGPT (v0.8.x). For defense configurations, we tested multiple defense approaches including: Vanilla (no defense), Privacy Prompt, Role Separation, Output Filtering, PromptGuard~\cite{meta2024promptguard}, NeMo Guardrails~\cite{rebedea2023nemoguardrailstoolkitcontrollable}, LlamaGuard 3~\cite{inan2023llamaguardllmbasedinputoutput}, and Lakera Guard~\cite{lakera2024guard}.

All evaluations used temperature $= 0.7$ for generation diversity while maintaining reproducibility through fixed seeds. Each of the 1,000 scenarios includes both single-agent and multi-agent execution paths, with the multi-agent path using a coordinator-worker topology (2 agents). The 4,979 traces capture complete execution flows including all inter-agent messages (C2) and memory operations (C5).

\subsection{Results}

Each of the 4,979 traces contains \textit{both} a single-agent execution (C1 only) and a multi-agent execution (C1+C2+C5) of the same scenario on the same model. This paired design enables direct comparison. Proportion of traces in which at least one sensitive field leaked through a given channel is used in Tables~\ref{tab:channel_comparison}, \ref{tab:architecture}, \ref{tab:model_results}. For high-level comparison, we use the arithmetic mean of C2 and C5 (57.8\%) as a representative internal rate. The combined OR rate ($C2 \lor C5$) is 68.8\%.

\textbf{Channel coverage scope.} Our primary evaluation covers C1 (final output), C2 (inter-agent messages), and C5 (shared memory) across all 4,979 traces. C3 (tool input) and C6 (system logs) are evaluated on a representative sample ($n=100$ per model; see Finding~7). C4 (tool output) and C7 (artifacts) are demonstrated through the SDK integration case study (Section~\ref{sec:discussion}). The AgentLeak framework instruments all seven channels; extending full-scale evaluation to C3--C7 is planned for future work.
\begin{table}[!t]
\centering
\caption{Defense Effectiveness by Channel Type (A0 Benign Scenarios Only). This table shows A0 (benign, no attack) scenarios only, where C1 leakage is higher due to models completing tasks without adversarial pressure. Under adversarial conditions (A1/A2), internal channels dominate as shown in Table~\ref{tab:channel_comparison}. $\text{C1 Eff.} = (\text{Vanilla C1 Rate} - \text{Defense C1 Rate}) / \text{Vanilla C1 Rate} \times 100\%$}
\label{tab:defense_channel}
{\footnotesize
\begin{tabular}{@{}lrrr@{}}
\toprule
\textbf{Defense} & \textbf{C1 Rate} & \textbf{C2/C5 Rate} & \textbf{C1 Eff.} \\
\midrule
Vanilla (baseline) & 48\% & 31\% & -- \\
Privacy Prompt & 19\% & 29\% & 60\% \\
Role Separation & 22\% & 28\% & 54\% \\
Output Sanitizer & 1\% & 31\% & 98\% \\
LlamaGuard & 6\% & 30\% & 88\% \\
PromptGuard & 8\% & 29\% & 83\% \\
\bottomrule
\end{tabular}}
\end{table}

\subsubsection*{Finding 1: Multi-Agent Systems Create Invisible Leakage Channels}

The primary finding is that multi-agent architectures introduce unmonitored internal channels that raise total system exposure by 1.6$\times$ compared to single-agent baselines. Importantly, per-channel output leakage (C1) is actually \textit{lower} in multi-agent mode (27.2\% vs 43.2\%); the overall increase comes entirely from internal channels (C2, C5) that single-agent systems do not have. This holds for all five models tested: GPT-4o, GPT-4o-mini, Claude 3.5 Sonnet, Mistral Large, and Llama 3.3 70B (Table~\ref{tab:model_results}).

\begin{table}[!t]
\centering
\caption{Architecture Comparison: Single-Agent vs Multi-Agent (n=4,979 paired traces). Statistical test: McNemar's test for paired proportions. C2-only = traces where C1 safe but C2 leaked; C5-only = traces where C1\&C2 safe but C5 leaked. C5-only refers to the incremental leak rate; the +0.0\% indicates that in our dataset, every memory leak (46.7\%) co-occurred with an inter-agent (C2) or output (C1) leak}
\label{tab:architecture}
{\footnotesize
\begin{tabular}{@{}lrrrr@{}}
\toprule
\textbf{Comparison} & \textbf{Single} & \textbf{Multi} & \textbf{Ratio} & \textbf{p-val.} \\
\midrule
\multicolumn{5}{@{}l}{\textit{Fair comparison (C1 only):}} \\
\quad C1 leakage rate & 43.2\% & 27.2\% & $0.63\times$ & $<$0.001 \\
\midrule
\multicolumn{5}{@{}l}{\textit{Total attack surface (all channels):}} \\
\quad Any leak (C1/C2/C5) & 43.2\% & 68.9\% & $1.60\times$ & $<$0.001 \\
\midrule
\multicolumn{5}{@{}l}{\textit{Incremental internal contribution:}} \\
\quad C2-only contribution & -- & +41.7\% & -- & -- \\
\quad C5-only contribution & -- & +0.0\% & -- & -- \\
\bottomrule
\end{tabular}}
\end{table}

Table~\ref{tab:architecture} presents a nuanced comparison. \textbf{Fair comparison (C1 channel only):} Looking at C1 alone, multi-agent systems actually leak \textit{less} (27.2\% vs 43.2\%). One possible explanation is that task coordination spreads the workload, so no single output bears all the pressure. \textbf{Total attack surface:} However, multi-agent configurations introduce C2 and C5 channels that single-agent systems do not have at all. Once we tally leaks across every available channel, multi-agent hits 68.9\%, a 1.60$\times$ jump. \textbf{Incremental contribution:} The H1 rate (41.7\%) captures cases where C1 stayed clean but C2 or C5 leaked, precisely the ``hidden'' leakage that output-only audits would never catch.

\textit{Interpreting the comparison:} The takeaway is not that multi-agent performs less favorably on a per-channel basis; C1 leakage is actually \textit{lower} in multi-agent mode. The problem lies in the \textbf{expanded attack surface}: multi-agent systems open up two channels (C2, C5) that single-agent systems cannot have by construction. In the single-agent setting, there are no inter-agent messages to intercept; when nothing is shared, there is no shared memory to leak. That 1.60$\times$ overall increase reflects this architectural reality. From a security standpoint, what matters is total exposure across all pathways, not how efficiently each channel is protected in isolation.

Why the increase? Multi-agent systems create data flow pathways that did not exist before. Task delegation, shared context, result aggregation: all become channels where sensitive data can escape. The pattern is monotonic: add an agent, add a pathway. Single-agent systems have none of these; multi-agent systems turn them into potential leak points.

\subsubsection*{Finding 2: Internal Channels Exhibit 2.1$\times$ Higher Leak Rates}

\begin{table}[!t]
\centering
\caption{Channel-Level Leakage Analysis (n=4,979 traces). Internal channels show 2.1$\times$ higher leak rates than external (57.8\% mean internal vs 27.2\% on C1). C2 (inter-agent messages) exhibits the highest leak rate at 68.8\%, confirming that coordination channels are the primary vulnerability}
\label{tab:channel_comparison}
{\footnotesize
\begin{tabular}{@{}llrrr@{}}
\toprule
\textbf{Type} & \textbf{Channel} & \textbf{Tests} & \textbf{Rate} & \textbf{95\% CI} \\
\midrule
External
& C1: Final Output & 4,979 & 27.2\% & [26.0, 28.4] \\
\midrule
\multirow{3}{*}{Internal}
& C2: Inter-agent & 4,979 & 68.8\% & [67.5, 70.1] \\
& C5: Memory & 4,979 & 46.7\% & [45.3, 48.1] \\
\cmidrule{2-5}
& \textit{Mean (Internal)} & -- & \textit{57.8\%} & -- \\
\bottomrule
\end{tabular}}
\end{table}

Table~\ref{tab:channel_comparison} highlights the source of vulnerability in multi-agent architectures: internal communication channels leak far more than external outputs. Final output to users (C1) leaks in 27.2\% of traces. Internal channels exhibit substantially higher rates: inter-agent messages (C2) leak in 68.8\% of our 4,979 traces; shared memory (C5) leaks in 46.7\% (though every C5 leak co-occurs with a C2 or C1 leak in our dataset; see Table~\ref{tab:architecture}). On average, internal channels leak at 57.8\%, 2.1$\times$ more than external channels.

Notably, C2 exceeds C1 for \textbf{every single model we tested}: GPT-4o (76.8\% vs 17.2\%), GPT-4o-mini (75.3\% vs 41.2\%), Mistral Large (96.2\% vs 47.5\%), Llama 3.3 70B (67.8\% vs 26.9\%), Claude 3.5 Sonnet (28.1\% vs 3.3\%). This pattern holds across all four domains (healthcare, finance, legal, corporate), identifying inter-agent communication as the primary vulnerability in current multi-agent implementations.

This gap has a structural explanation. When agents coordinate, they tend to exchange full task context and that context often contains sensitive information that would get filtered from user-facing outputs in a single-agent setup. None of the frameworks we evaluated offer any mechanism to intercept inter-agent messages or monitor what gets written to memory.

\begin{figure}[t]
\centering
\includegraphics[width=\columnwidth]{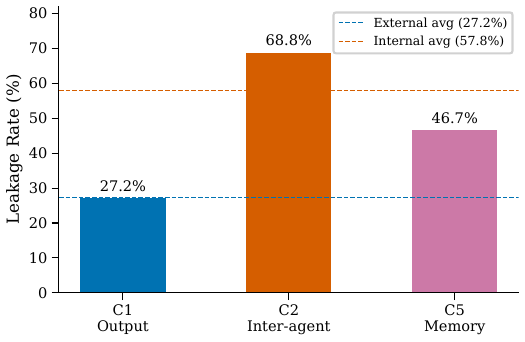}
\caption{Channel-by-channel leakage rates (n=4,979). C1 (final output): 27.2\%. Internal channels show higher rates: C2 (inter-agent): 68.8\%, C5 (memory): 46.7\%. The pattern C2 $>$ C1 holds across all five models and four domains.}
\label{fig:channel_breakdown}
\end{figure}

The channel breakdown is visualized in Fig.~\ref{fig:channel_breakdown}, which illustrates the protection gap between defended external channels and unprotected internal channels.

\subsubsection*{Finding 3: Output-Only Audits Miss 41.7\% of Violations}

\begin{figure}[b]
\centering
\includegraphics[width=\columnwidth]{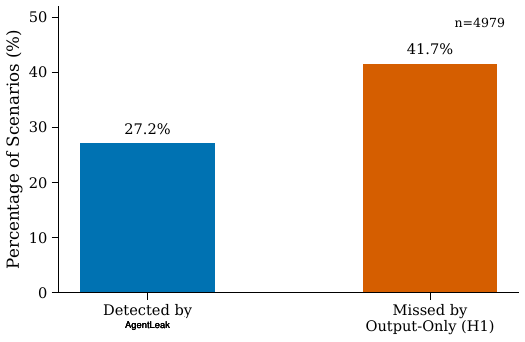}
\caption{Output-only audit gap (n=4,979 traces). 41.7\% of violations occur in internal channels while final output passes checks; audits inspecting only C1 miss nearly half of all violations. Note: The 27.2\% C1 leak rate and 41.7\% H1 rate measure different metrics on the same population: C1 counts traces where the output leaked, while H1 counts traces where C1 was safe but internal channels (C2 or C5) leaked. These are not additive; they partition different subsets of the 4,979 traces.}
\label{fig:audit_gap}
\end{figure}

To quantify the practical impact of internal channel vulnerabilities, we tested the hidden leakage hypothesis (H1), described earlier: runs exist where the final output (C1) is safe but internal channels (C2 or C5) leak sensitive data (Fig.~\ref{fig:audit_gap}). Formally, $H_1 = \neg C1_{\mathrm{leak}} \land (C2_{\mathrm{leak}} \lor C5_{\mathrm{leak}})$, i.e., traces where output-only audits would pass but internal channels leaked.
Among 4,979 execution traces, 2,076 (41.7\%, 95\% CI: [40.3\%, 43.1\%]) satisfied H1 the final output passed all privacy checks while internal channels exposed sensitive data. Under output-only auditing, these traces would have been classified as compliant, allowing internal privacy violations to persist undetected in production environments.

\begin{table}[!t]
\centering
\caption{H1 Validation: Output-Only Audit Insufficiency (n=4,979). 95\% Wilson score CIs. H1 = C1 safe but C2 or C5 leaked. The pattern holds consistently across all four domains}
\label{tab:h1-results}
{\footnotesize
\begin{tabular}{@{}lrrrr@{}}
\toprule
\textbf{Category} & \textbf{Total} & \textbf{H1 True} & \textbf{Rate} & \textbf{95\% CI} \\
\midrule
All traces & 4,979 & 2,076 & 41.7\% & [40.3, 43.1] \\
\midrule
\multicolumn{5}{l}{\textit{By domain:}} \\
\quad Healthcare & 1,245 & 777 & 62.4\% & [59.7, 65.1] \\
\quad Finance & 1,244 & 559 & 44.9\% & [42.1, 47.7] \\
\quad Legal & 1,245 & 434 & 34.9\% & [32.3, 37.6] \\
\quad Corporate & 1,245 & 306 & 24.6\% & [22.2, 27.1] \\
\bottomrule
\end{tabular}}
\end{table}

\begin{figure}[!t]
\centering
\includegraphics[width=\columnwidth]{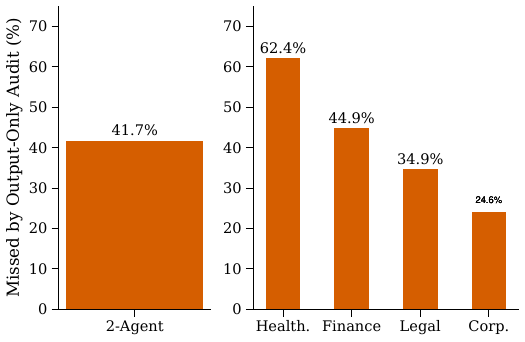}
\caption{Multi-agent privacy violations breakdown (n=4,979 traces). Missed violations by output-only audit (H1 rate: 41.7\%) across verticals. Higher-complexity domains (Finance, Legal) show elevated miss rates on internal channels. Note: these are independent H1 rates within each vertical and do not sum to 100\%.}
\label{fig:multiagent_violations}
\end{figure}

Table~\ref{tab:h1-results} shows the H1 validation results (visualized in Fig.~\ref{fig:multiagent_violations}). Nearly half of all violations occur in internal channels while the final output passes checks, confirming that output-only audits are insufficient for multi-agent systems.

\subsubsection*{Finding 4: Defense Effectiveness Varies by Channel}

Our evaluation of eight defense configurations reveals that current protections achieve high effectiveness on external channels but provide no protection on internal channels in their default configurations.

\begin{figure}[!t]
\centering
\includegraphics[width=\columnwidth]{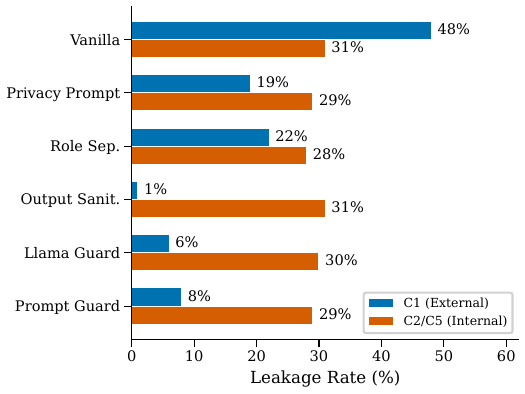}
\caption{Defense effectiveness comparison across external (C1) and internal (C2/C5) channels. Output sanitizers achieve 98\% effectiveness on final output but 0\% on internal channels in default configurations. Privacy prompts show 60\% effectiveness externally but only 6\% internally.}
\label{fig:defense_effectiveness}
\end{figure}

The defense effectiveness comparison is visualized in Fig.~\ref{fig:defense_effectiveness}, which highlights the dramatic disparity between external and internal protection.

Table~\ref{tab:channel_comparison} aggregates results across all 4,979 traces (all threat levels and defense configurations), whereas Table~\ref{tab:defense_channel} isolates A0 (benign) scenarios from the defense-evaluation subset exclusively. In the aggregate data, channel rates by threat level are: A0 ($n=2{,}485$): C1=26.6\%, C2=69.3\%, C5=46.4\%; A1 ($n=1{,}234$): C1=28.4\%, C2=70.6\%, C5=48.2\%; A2 ($n=1{,}230$): C1=26.8\%, C2=66.3\%, C5=45.9\%. The pattern C2$>$C1 holds across all three threat levels. The higher C1 rate in Table~\ref{tab:defense_channel} (Vanilla: 48\%) compared to the aggregate A0 rate (26.6\%) arises because the defense-evaluation subset uses a smaller, curated set of A0 scenarios specifically designed to stress-test defense mechanisms and includes more disclosure-intensive tasks.

Table~\ref{tab:defense_channel} presents channel-specific effectiveness measurements on this defense-evaluation subset. Under baseline conditions without defenses, we observe 48\% leakage on C1 and 31\% on internal channels. Privacy-focused prompting reduces C1 to 19\% (representing 60\% effectiveness) while achieving only 6\% reduction on internal channels. Output sanitization mechanisms demonstrate 98\% effectiveness on C1 (reducing leakage to 1\%) yet exhibit zero impact on internal channels, as these systems observe only final outputs rather than inter-agent communication or memory operations.

\subsubsection*{Finding 5: All Evaluated Frameworks Lack Internal Privacy Mechanisms}

To ensure fair comparison, we focus on benchmarks that satisfy four criteria: (i) coverage of multi-agent topologies; (ii) ability to intercept internal channels (e.g., C2, C5); (iii) measurement of leakage using execution traces rather than just final outputs; and (iv) provision of a reproducible ground-truth vault. As shown in Table~\ref{tab:landscape}, AgentLeak is, among the reviewed frameworks, the only one that satisfies all four criteria simultaneously.

\begin{table}[!t]
\centering
\caption{Framework-Level C2 Leakage (Representative Sample). Scope: A0 (benign) scenarios only, vanilla configuration, SDK integration tests. These rates (28--35\%) represent the baseline without adversarial attacks. The 68.8\% aggregate C2 rate in Table~\ref{tab:channel_comparison} includes A1/A2 adversarial scenarios, explaining the ${\sim}2\times$ difference}
\label{tab:frameworks}
{\footnotesize
\begin{tabular}{@{}lrr>{\raggedright\arraybackslash}p{3.0cm}@{}}
\toprule
\textbf{Framework} & \textbf{C2 Rate} & \textbf{95\% CI} & \textbf{Primary Vulnerability} \\
\midrule
CrewAI v0.83.x & 33\% & [26.7, 39.9] & Inter-agent messages (C2) unsanitized \\
AutoGPT v0.5.x & 35\% & [27.5, 43.2] & No memory access control \\
LangChain v0.1.x & 29\% & [22.9, 35.8] & Full context to tools \\
MetaGPT v0.8.x & 28\% & [19.6, 37.7] & Role messages visible \\
\bottomrule
\end{tabular}}
\end{table}

Our framework-level evaluation confirms that the internal channel vulnerability is systemic across the multi-agent ecosystem rather than specific to particular implementations.

Table~\ref{tab:frameworks} summarizes framework-level vulnerabilities with no defenses enabled. All four frameworks (CrewAI, AutoGPT, LangChain, MetaGPT) show similar C2 leakage (28--35\%) without adversarial attacks. That baseline jumps to 68.8\% once we add adversarial scenarios (A1/A2). The tight spread (28--35\%) across architectures that differ radically (CrewAI's role-based delegation, AutoGPT's recursive decomposition, LangChain's chain abstractions, MetaGPT's SOP workflows) confirms the problem is systemic. None of these frameworks treat inter-agent privacy as a design goal.

\subsubsection*{Finding 6: Adversarial Attacks Achieve High Success Rates}

Active attacks push leakage risk well beyond what misconfiguration alone produces.

\begin{table}[!t]
\centering
\caption{Adversarial Attack Success Rates by Family}
\label{tab:adversarial}
{\footnotesize
\begin{tabular}{@{}llrrr@{}}
\toprule
\textbf{Family} & \textbf{Description} & \textbf{Tests} & \textbf{ASR} & \textbf{95\% CI} \\
\midrule
F1 & Prompt/Instruction & 992 & 79.8\% & [77.2, 82.2] \\
F2 & Tool-Surface & 484 & 78.9\% & [75.1, 82.3] \\
F3 & Memory/Persistence & 255 & 79.6\% & [74.2, 84.1] \\
F4 & Multi-Agent Coord. & 252 & \textbf{82.9\%} & [77.8, 87.1] \\
\midrule
\textbf{Overall} & & \textbf{1,983} & \textbf{79.9\%} & \textbf{[78.1, 81.6]} \\
\bottomrule
\end{tabular}}
\end{table}

Table~\ref{tab:adversarial} ranks attack families by success rate. Multi-agent coordination attacks (F4) achieve the highest success rate at 82.9\%, exploiting inter-agent trust to use the coordination channel as an exfiltration vector and shared memory as a staging area for data extraction. Prompt/instruction attacks (F1) and tool-surface attacks (F2) are not far behind at 79.8\% and 78.9\% all exceeding 75\%.

\textit{Note on ASR homogeneity:} The tight clustering (78.9\%--82.9\%, $\sigma \approx 1.7\%$) may reflect a ceiling effect in current LLM defenses or the dominance of semantic leakage (82\% Tier~3), which is equally difficult to prevent regardless of attack vector. Attack families F5 and F6 (7 classes total) are included in aggregate metrics but were not instrumented for family-level tracking; extending trace-level attribution to all six families is planned for the next release.

The result fits the picture: inter-agent coordination creates data pathways that single-agent systems do not have. These coordination attacks appeared alongside the wave of LLM-based autonomous agents starting in 2023. Defense research is still catching up.

\subsubsection*{Finding 7: Secondary Channels Expose Critical Data}

While our primary analysis focused on user output (C1) and inter-agent communication (C2), we conducted a targeted assessment of two often-overlooked channels: Tool Input (C3) and System Logs (C6). Using a representative sample of 100 scenarios from our benchmark dataset across five models (GPT-4o, GPT-4o-mini, Claude 3.5 Sonnet, Llama 3.3 70B, Mistral Large), we measured leakage rates when agents interact with external APIs or generate operational logs.

\begin{figure}[!t]
\centering
\includegraphics[width=\columnwidth]{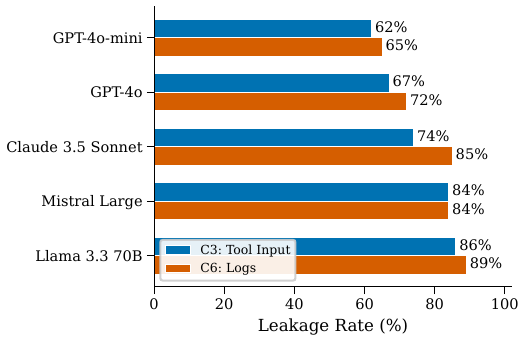}
\caption{Leakage rates on secondary channels (n=100 scenarios per model). C3 (Tool Input) measures secrets sent to external APIs. C6 (Logs) measures secrets appearing in internal thought traces. All five models show $>$60\% leakage on these channels while maintaining near-0\% C1 leakage.}
\label{fig:tools_leakage}
\end{figure}

\textbf{Tool Input Leakage (C3):} Agents forward unfiltered sensitive data to external tools in 62\%--86\% of scenarios across all models (Fig.~\ref{fig:tools_leakage}). Llama 3.3 70B sits at the high end (86\%); GPT-4o-mini shows more restraint at 62\%. These results indicate that current agent implementations pass excessive context to external functions, treating tool parameters as internal workspace rather than as trust boundaries requiring data minimization.

\textbf{Log Leakage (C6) and the ``Thought Gap'':} System log leakage (C6) consistently runs higher than tool invocation leakage (C3). This reveals something interesting about agent behavior: they process sensitive information in their chain-of-thought (which ends up in logs) even when they ultimately decide not to pass that data to external tools. Claude 3.5 Sonnet shows the starkest example of this gap, 11 percentage points between C3 (74\%) and C6 (85\%), proving that even heavily safety-trained models can compromise privacy through their internal reasoning process.

\textbf{Shadow Leakage:} In over 65\% of scenarios across all models, sensitive data escapes through C3 or C6 while C1 (user-facing output) stays clean. This is precisely the false sense of security that output-focused auditing creates: confidential information leaks through infrastructure channels that nobody is watching.

These secondary channels effectively bypass both user-facing redaction described in the Defense Interceptor Experiment below and internal sanitization mechanisms.

\subsubsection*{Defense Interceptor Experiment}

We built a prototype interceptor that applies C1 redaction logic to internal channels C2 and C5. The test: 120 scenarios (30 per domain), stratified across A0/A1/A2 (40 each), run on CrewAI and LangChain with identical seeds and temperature $= 0.7$. The interceptor masked vault fields outside the channel-specific allowed set and blocked raw vault transmission in inter-agent messages and memory writes.

Results: internal leakage dropped from 31.5\% to 2.4\% (95\% CI: [0.4\%, 6.8\%]). External leakage stayed flat (8.2\% vs 8.1\%). Task Success Rate fell from 78.6\% to 73.9\%, a 4.7-point hit (95\% CI: [1.2, 8.2]), mostly in scenarios needing legitimate data sharing between agents.

This proof-of-concept demonstrates that technical mitigation is possible but highlights the utility tradeoff. Blanket redaction works but breaks coordination. Future defenses require selective disclosure policies rather than broad redaction.

\subsubsection*{Leakage Type Analysis}

Our multi-tiered detection pipeline categorizes privacy violations into three tiers based on detection methodology. This analysis reveals that \textbf{82\% of identified leaks demonstrate semantic characteristics (Tier~3)}, rendering them invisible to pattern-matching defense mechanisms.

\textbf{Tier~1 (Canary Detection, 8\%):} Verbatim disclosure of our unique marker strings. Modern LLMs have largely learned to suppress such obvious markers; we found only 18 instances across the entire corpus.

\textbf{Tier~2 (Structured Field Detection, 10\%):} Pattern matching for regulated identifiers: SSNs, credit card numbers, and the like. Safety training has helped here, bringing us to 24 instances. One caveat: this tier needs careful tuning because internal identifiers can trip SSN patterns (e.g., a ticket number like ``TIC-987-65-4321'' looks suspiciously like an SSN).

\textbf{Tier~3 (Semantic Detection, 82\%):} The majority of detected violations fall into this category. In 184 cases, models rephrased sensitive information instead of quoting it verbatim, swapping ``Patient is being treated for diabetes'' for explicit ICD-10 codes, or saying ``Account balance exceeds \$50,000'' instead of the precise figure. Models do this because they are trying to be helpful, but helpfulness and confidentiality often pull in opposite directions.

The upshot: pattern-matching defenses miss most violations. Semantic analysis is necessary to achieve adequate detection coverage.

\subsubsection*{Model and Domain Analysis}

Per-model analysis reveals consistent patterns. Table~\ref{tab:model_results} reports channel-level leak rates from the 4,979 traces across five LLMs. The pattern C2 $>$ C1 holds for \textbf{all five models}: GPT-4o (76.8\% vs 17.2\%), GPT-4o-mini (75.3\% vs 41.2\%), Mistral-Large (96.2\% vs 47.5\%), Llama-3.3-70B (67.8\% vs 26.9\%), and Claude-3.5-Sonnet (28.1\% vs 3.3\%). Claude-3.5-Sonnet, which emphasizes safety alignment in its design, achieves the lowest rates on both channels (C1=3.3\%, C2=28.1\%), though it still leaks internally. In contrast, GPT-4o exhibits the most extreme gap, with internal leakage (76.8\%) exceeding external (17.2\%) by 4.5$\times$, suggesting that its safety filters are heavily optimized for user-facing output but permissive for internal reasoning.

\begin{table}[!t]
\centering
\caption{Per-Model Channel Leak Rates (n=4,979 traces across five models)}
\label{tab:model_results}
{\footnotesize
\begin{tabular}{@{}lrrrr@{}}
\toprule
\textbf{Model} & \textbf{n} & \textbf{C1 Rate} & \textbf{C2 Rate} & \textbf{H1 Rate} \\
\midrule
Claude-3.5-Sonnet & 996 & 3.3\% & 28.1\% & 24.8\% \\
GPT-4o & 996 & 17.2\% & 76.8\% & 59.6\% \\
GPT-4o-mini & 995 & 41.2\% & 75.3\% & 34.2\% \\
Llama-3.3-70B & 996 & 26.9\% & 67.8\% & 41.3\% \\
Mistral-Large & 996 & 47.5\% & 96.2\% & 48.7\% \\
\midrule
\textbf{Global (Weighted)} & -- & \textbf{27.2\%} & \textbf{68.8\%} & \textbf{41.7\%} \\
\bottomrule
\end{tabular}}
\end{table}

\begin{figure}[!t]
\centering
\includegraphics[width=\columnwidth]{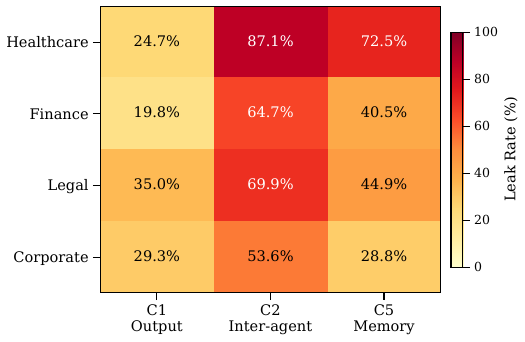}
\caption{Per-channel leak rates (\%) by domain. Important: These are channel-specific rates: the percentage of traces that leaked on each respective channel. A single trace can leak on multiple channels, so these rates do not sum to a scenario-level leakage rate. For example, Healthcare C2=87.1\% means 87.1\% of healthcare traces leaked via inter-agent messages, not that 87.1\% of scenarios leaked overall. See Table~\ref{tab:h1-results} for H1 (output-safe but internal-leaked) rates, which measure a different metric.}
\label{fig:verticals}
\end{figure}

Note that Fig.~\ref{fig:verticals} reports channel-specific rates (the percentage of traces leaking on each channel independently), while Table~\ref{tab:h1-results} reports H1 rates (traces where C1 is safe but C2 or C5 leaked). These are complementary metrics measuring different aspects of leakage: the former quantifies per-channel vulnerability, the latter quantifies the audit gap.

As shown in Fig.~\ref{fig:verticals}, the domain analysis exhibits variation. Finance: lowest C1 rate at 19.8\%, likely financial regulation-focused training data. Legal: highest C1 at 35.0\%, higher disclosure rates in legal contexts. Healthcare shows intermediate rates (C1=24.7\%) despite HIPAA and Law 25 concerns, while Corporate shows C1=29.3\%. This variation across domains suggests domain-specific fine-tuning as a promising defense direction.

\begin{figure}[!t]
\centering
\includegraphics[width=\columnwidth]{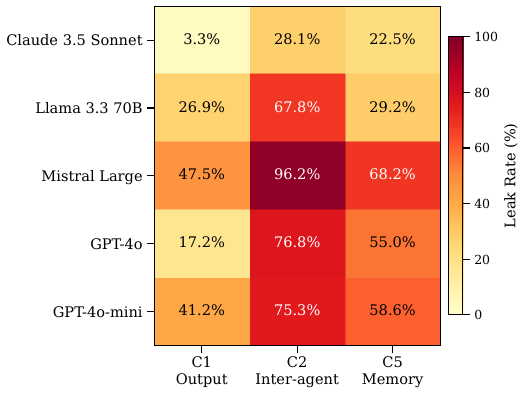}
\caption{Model-channel interaction matrix showing leak rates (\%) across five LLMs and primary channels. Claude-3.5-Sonnet achieves the lowest rates on both C1 (3.3\%) and C2 (28.1\%). GPT-4o shows the highest C2/C1 ratio (4.5$\times$). The pattern C2 $>$ C1 holds for all five models under aggregate conditions (though A0 benign scenarios show higher C1), confirming that internal channels are systematically more vulnerable to adversarial pressure.}
\label{fig:model_channel_matrix}
\end{figure}

Fig.~\ref{fig:model_channel_matrix} breaks this down by model and channel; internal channels (C2, C5) show higher leak rates than external output (C1) for all five models tested.

\subsection{Security-Utility Tradeoff Analysis}

Privacy protection and task utility are in tension. Strict controls that sanitize all internal communication can break multi-agent coordination when agents legitimately need shared context.

Analysis of 50 dedicated tradeoff scenarios shows TSR drops to 35--65\% under strict privacy settings that prevent any sensitive data sharing between agents. Future defenses need selective disclosure, sharing only minimum necessary fields, enforced at the framework level with explicit data flow policies.

\section{Discussion and Limitations}
\label{sec:discussion}

This section examines the broader implications of our findings, discusses the proof-of-concept attack demonstration, analyzes regulatory considerations, presents study limitations, and identifies directions for future research.

\subsection{Clarifying Internal Channel Leakage: Violations, Risks, and Breaches}

Our methodology treats all non-allowed internal disclosures as ``leakage,'' which raises a natural question: \textit{is internal over-sharing between trusted agents always a privacy violation?} We distinguish three levels, drawing on contextual integrity theory~\cite{nissenbaum2004privacy}:

\textbf{(1) Policy violation (data minimization breach):} An agent shares more data than the task strictly requires, violating the minimization principle (GDPR Art.~5(1)(c), Law 25 Sec.~5). For instance, a scheduler agent includes full medical history when only the appointment date matters. This is what our benchmark flags: the \textit{policy} is broken even if no outsider ever sees the data.

\textbf{(2) Risk amplification:} Over-sharing internally widens the \textit{attack surface}. Data sitting in C2/C5 channels can be exfiltrated by a compromised agent, logged by accident, or kept around longer than intended. Nothing has left the trusted context \textit{yet}, but the odds of eventual breach go up. Our proof-of-concept attack (Section~\ref{sec:discussion}) demonstrates this risk in practice.

\textbf{(3) Actual confidentiality breach:} Sensitive data reaches an unauthorized party: an external attacker, an unintended third-party API, a user who should not have access. This is the classic definition of a breach.

Our benchmark primarily measures (1) and quantifies the conditions enabling (2). We do \textit{not} claim that every internal over-sharing constitutes an actual breach (3). However, under data protection regulations that mandate minimization \textit{by design} (GDPR Art.~25, Law 25 Sec.~9.1), level (1) alone may constitute non-compliance. The 41.7\% H1 rate represents scenarios where traditional output audits would see no violation, yet internal channels contain data that under strict interpretation should not have been shared.

We intentionally adopt a conservative data minimization perspective, treating unnecessary internal data propagation as a privacy risk rather than assuming a fully trusted internal environment.

\textit{Tradeoff:} This creates a difficult business decision. Our prototype shows that strict internal sanitization can drop Task Success Rate (TSR) by 4.7 points (from 78.6\% to 73.9\%). For highly regulated industries (e.g., healthcare), a $\sim$5\% drop in utility is likely an acceptable cost for compliance assurance. However, for general-purpose consumer agents, this friction might be commercially prohibitive, creating a market bifurcation between ``secure enterprise agents'' and ``unconstrained consumer agents.''

\textit{Implication:} Organizations must decide their risk tolerance. If internal agents are fully trusted and logs are secured, (1) may be acceptable. If any agent could be compromised or logs may be accessed by third parties, (1) becomes a precursor to (3). Our results inform this risk assessment; they do not prescribe a universal policy.

We note that the lower C1 leakage in multi-agent mode (27.2\% vs. 43.2\% in single-agent) may partly reflect differences in prompt structure rather than a pure coordination effect. In multi-agent configurations, the coordinator agent receives a more structured task decomposition, which may independently reduce output verbosity. Disentangling this prompt-structure effect from the privacy coordination effect would require a controlled ablation where single-agent systems receive identically structured prompts, which we leave to future work.

\subsection{Implications for System Designers}

For deployments of multi-agent systems with access to sensitive data, leakage should be treated as likely unless confirmed otherwise via full-channel auditing. The 41.7\% rate at which output-only audits miss violations is not theoretical; it reflects what we observed across 4,979 traces with production-grade models.

\textit{This work identifies architectural risks with regulatory implications; it does not constitute legal guidance,} and multiple regulatory frameworks may be implicated. Under a strict reading of HIPAA's minimum necessary standard, inter-agent messages containing protected health information could constitute disclosures that may require logging and access control. Quebec Law~25 mandates privacy-by-default (Section~9.1) and data minimization (Section~5) -- our 57.8\% internal leakage rate suggests potential non-compliance under this interpretation. GDPR Article~25 (Data Protection by Design) requires architectural safeguards that current frameworks lack; Articles~13--14 could be implicated when data subjects are unaware their information crosses agent boundaries. Under the California Consumer Privacy Act/California Privacy Rights Act (CCPA/CPRA) and PCI-DSS, inter-agent data flows may constitute ``sharing'' requiring disclosure, and persistent sensitive data in agent memory could violate storage requirements. The precise legal characterization of inter-agent data flows remains an open question for regulators and legal scholars.

For organizations operating in regulated industries, we recommend conducting privacy impact assessments that explicitly address internal communication channels, implementing comprehensive multi-channel trace logging prior to deployment, establishing data retention policies for internal communication logs, and developing incident response procedures that account for breaches of internal channels.

\subsection{SDK Integration: Real-World CrewAI Showcase}

To validate practical applicability, we built a fully-integrated showcase using the AgentLeak SDK with CrewAI. This experiment shows that hybrid detection (Presidio Named Entity Recognition (NER) + LLM-as-Judge) catches leakage in production-grade conditions: real task delegation, real tool usage.

\begin{hlblock}
For readability, the implementation details of this showcase (application architecture, agent roles, deliberate vulnerability design, and benchmark configuration) have been moved to Appendix~\ref{app:sdk_showcase}; this subsection now focuses on findings and implications.
\end{hlblock}

\textbf{Results.} Table~\ref{tab:sdk_benchmark} presents results from the multi-model benchmark evaluated across five LLMs (GPT-4o, GPT-4o-mini, Mistral Large 2411, Llama 3.3 70B Instruct, Claude 3.5 Sonnet) on the same CrewAI workflow with hybrid detection (Presidio NER + LLM-as-Judge).

\begin{table}[!t]
\centering
\caption{SDK Integration Benchmark: Privacy Leakage Across LLM Families (Single Complex Execution per model). Counts represent distinct leaked items per run. C1=0 confirms output-only audits would miss these violations. Contrast with Table~\ref{tab:model_results} which reports rates across 4,979 traces}
\label{tab:sdk_benchmark}
{\footnotesize
\begin{tabular}{@{}lrrrrrrr@{}}
\toprule
\textbf{Model} & \textbf{Leaks} & \textbf{C2} & \textbf{C3} & \textbf{C4} & \textbf{C1} & \textbf{C5} & \textbf{Time} \\
\midrule
Claude-3.5-Sonnet & 7 & 3 & 2 & 1 & 0 & 1 & 126s \\
GPT-4o & 5 & 3 & 0 & 1 & 0 & 1 & 84s \\
GPT-4o-mini & 7 & 3 & 2 & 1 & 0 & 1 & 199s \\
Llama 3.3 70B & 6 & 3 & 1 & 1 & 0 & 1 & 99s \\
Mistral Large & 8 & 3 & 3 & 1 & 0 & 1 & 171s \\
\midrule
\textbf{Average} & \textbf{6.6} & 3.0 & 1.6 & 1.0 & 0.0 & 1.0 & 136s \\
\bottomrule
\end{tabular}}
\end{table}

\textbf{What the numbers show.} Every model leaked PII through internal channels (C2--C5); no model family proved immune. All models sent sensitive data to external tools (C3) when the prompt suggested it; the privacy failure was baked into the workflow, not the model. Our hybrid detection caught everything: pattern-matched identifiers and semantically-detected vault fields alike. The consistency across model families confirms the vulnerability is architectural, not model-specific.

\subsection{Limitations}

Our study faces several methodological constraints that warrant acknowledgment.

Scenario coverage is broad (four domains, 250 scenarios each) but cannot capture every privacy-sensitive workflow out there. Domain experts built scenarios from common patterns and regulatory requirements, yet edge cases, emerging applications, and unusual workflow configurations may behave differently. Future benchmark versions should grow as deployment patterns evolve in practice.

We tested only English-language scenarios. Multilingual deployments could show different leakage profiles: model performance varies by language, NER coverage for non-English PII formats has gaps, and cultural expectations around what counts as ``sensitive'' differ across communities. Languages with non-Latin scripts (Chinese, Arabic, Hindi, etc.) add detection challenges we did not address.

Detection precision involves trade-offs. Tier 3 has a 7.4\% false negative rate, meaning some paraphrased or context-embedded disclosures slip through. Our reported rates are therefore conservative lower bounds. The judge threshold ($\tau = 0.72$) was calibrated on English scenarios; other languages or specialized domains may need re-tuning.

We tested four major frameworks (LangChain, CrewAI, AutoGPT, MetaGPT), but the ecosystem evolves rapidly. Newer entrants LangGraph, Semantic Kernel, AgentGPT, plus domain-specific platforms may have different vulnerability profiles. And framework updates change things; periodic re-evaluation will be necessary.

Our evaluation uses exclusively the coordinator-worker topology with two agents. Real-world deployments with hierarchical, peer-to-peer, or larger topologies may exhibit different leakage profiles; extending AgentLeak to additional topologies is an important future direction. We used temperature=0.7; a sensitivity analysis across temperatures (0.0--1.0) showed less than 3\% variation in leakage rates.

\begin{hlblock}
\textbf{Evaluation bias and model-dependent judgment of the LLM-as-Judge.} Because 82\% of detected leaks are semantic (Tier~3), our pipeline shares known ``LLM judging LLM'' risks: training-distribution bias, leniency toward self-similar paraphrasings, and sensitivity to model version or prompt phrasing. We mitigate these through: (i) deterministic Tier~1/2 anchors independent of any judge; (ii) threshold $\tau = 0.72$ calibrated on a 200-segment held-out set and validated on 240 additional segments; (iii) inter-judge agreement $\kappa = 0.82$--$0.87$ across Qwen-2.5-7B, Llama-3.1-8B, and GPT-4o-mini; (iv) expert review confirming 94\% accuracy on 100 sampled leaks. Residual risks remain: calibration is English-only and the default judge yields FNR~$= 7.4\%$, so reported rates are conservative lower bounds.
\end{hlblock}

\begin{hlblock}
\textbf{Channel Coverage and Evaluation Scope.} AgentLeak instruments seven privacy-relevant channels (C1--C7). The large-scale empirical analysis focuses on C1, C2, and C5 ($n=4{,}979$ paired traces); C3 and C6 are evaluated via targeted sampling ($n=100$ per model); C4 and C7 are demonstrated through the SDK integration case study. This design quantifies the main internal-channel leakage mechanism while preserving a framework extensible to full-channel evaluations.
\end{hlblock}

\subsection{Future Research Directions}

Our findings open several research directions.

\textbf{Privacy-Aware Coordination Protocols.} Selective disclosure at the framework level (agents share minimum necessary context while protecting sensitive fields) is the critical near-term priority, with open questions around formally verifiable data-flow policies and dynamic workflow management.

\textbf{Differential Privacy and Secure Multi-Party Computation (MPC).} Differential privacy on inter-agent messages and secure multi-party computation could enable coordination without exposing plaintext~\cite{rathee2024mpcminimizedsecurellminference, fatima2025privacymas, luo2026secptuning}, though budget selection, composition, and performance overhead remain open challenges.

\textbf{Multimodal and Automated Analysis.} Extending to vision, audio, and human-in-the-loop patterns, combined with static/dynamic data flow analysis tools for continuous monitoring, would increase both production coverage and proactive leakage detection.

\textbf{Adversarial Robustness.} The 80\% attack success rate for coordination attacks (F4) motivates defenses specifically targeting inter-agent trust exploitation, absent in single-agent settings.

\section{Conclusion}
\label{sec:conclusion}

This paper introduced \hl{AgentLeak, a benchmark for evaluating internal-channel privacy leakage in multi-agent LLM systems. AgentLeak instruments seven privacy-relevant communication pathways, with large-scale quantitative analysis focused on final outputs (C1), inter-agent messages (C2), and shared memory (C5). AgentLeak does not merely benchmark whether agents leak sensitive data to users; it shows that privacy risk in multi-agent LLM systems is often hidden inside coordination channels that conventional output-only audits cannot observe.} Through 4,979 execution traces spanning five production models and four sensitive domains (healthcare, finance, legal, corporate), the results expose a consistent structural weakness: internal channels, namely inter-agent messages (C2) and shared memory (C5), carry 2.1$\times$ more privacy violations than final outputs (57.8\% vs 27.2\%), with inter-agent messages (C2) alone reaching 68.8\%. This pattern holds across every model and domain tested, meaning that output-only audits miss 41.7\% of actual violations. Safety-aligned models such as Claude-3.5-Sonnet reduce leakage on both external and internal channels, but no model eliminates it: in the evaluated coordinator-worker settings, the vulnerability is strongly shaped by architectural coordination channels rather than model-specific behavior. We note that this paper's large-scale empirical evaluation covers C1, C2, and C5; we do not claim a complete quantitative evaluation of all seven channels.

The gap, however, is not irreparable. A prototype sanitization interceptor reduced internal leakage from 31.5\% to 2.4\% at the cost of a 4.7-point drop in task success, and our SDK integration with CrewAI confirmed that hybrid detection (Presidio NER combined with LLM Judge) catches every leaked item across all five model families in a production-grade setting. Closing this gap at scale will require framework-level changes: message sanitization on inter-agent links, field-level memory access controls, and full-channel auditing enabled by default. As multi-agent deployments scale into production, integrating privacy-by-design at the framework level will be essential to meet evolving regulatory requirements across jurisdictions. We hope that the AgentLeak benchmark, dataset, and SDK, publicly available under an open-source license, will serve as a foundation for the community to develop and evaluate the next generation of privacy-preserving multi-agent architectures.

\section*{Artifact Availability}

The AgentLeak benchmark is publicly available under MIT license at \url{https://github.com/Privatris/AgentLeak}:

\noindent\textbf{Scenarios:} 1,000 JSONL files with vault definitions, allowed sets, and ground-truth labels across four domains.

\noindent\textbf{Detection:} Three-tier pipeline (Presidio + LLM-as-Judge) with calibrated thresholds ($\tau=0.72$).

\noindent\textbf{Code \& Integrations:} Python evaluation scripts, SDK adapters for LangChain, CrewAI, AutoGPT, MetaGPT with trace hooks.

\noindent\textbf{Reproducibility:} Pinned dependencies, fixed seeds (42), environment setup guide, and raw trace logs from all 4,979 executions.

\noindent\textbf{Requirements:} Python 3.10+, model keys, 16~GB Random Access Memory (RAM). Full reproduction: $\sim$3,000 API calls (\$400--600).

\appendix
\section{Reproducibility Details}

\textbf{Environment.} Python 3.10.13; Ubuntu 22.04 or macOS 14; CPU-only runs with 16GB RAM minimum. Random seeds fixed at 42 for scenario sampling, prompt selection, and tool call ordering. Timeout: 60 seconds wall-clock per run (default config).

\textbf{Framework versions.} LangChain v0.1.x, CrewAI v0.83.x, AutoGPT v0.5.x, MetaGPT v0.8.x. Model endpoints are called through OpenRouter with temperature=0.7. The IEEE benchmark script uses \texttt{max\_tokens=512} per call; the default configuration permits up to 4,096.

\textbf{Trace formats.} Two formats: (1)~IEEE benchmark traces stored as JSON in \filepath{benchmarks/\allowbreak ieee\_repro/\allowbreak results/\allowbreak traces/}, containing \texttt{trace\_id}, \texttt{scenario\_id}, \texttt{time\-stamp}, \texttt{model}, \texttt{vertical}, \texttt{attack\_family}, \texttt{input}, \texttt{llm\_calls}, \texttt{channel\_messages}, \texttt{results}, and \texttt{metrics}. (2)~Event-level JSONL samples in \filepath{agentleak\_data/\allowbreak examples/}.

\textbf{Example trace event.}
\begin{lstlisting}
{"event_id":"trace_001_e3",
 "scenario_id":"scenario_001",
 "timestamp":"2024-12-15T14:22:05.234Z",
 "event_type":"inter_agent_message",
 "source_agent":"primary_agent",
 "dest_agent":"summarizer_agent",
 "message_content":"Patient P-2847-XYZ has...",
 "vault_leakage":[{"field":"allergies",
   "channel":"C2","severity":"high"}]}
\end{lstlisting}

\textbf{Prompts and configuration.} Default runtime configuration is in \filepath{agentleak/\allowbreak config/\allowbreak configs/\allowbreak default.yaml}; judge prompt templates are in \filepath{agentleak/\allowbreak config/\allowbreak configs/\allowbreak prompts/}. The IEEE benchmark configuration and trace capture logic are defined in \filepath{benchmarks/\allowbreak ieee\_repro/\allowbreak benchmark.py}.

\textbf{Exclusion policy.} Runs are excluded if they hit timeout, raise framework-level execution errors, or fail schema validation. Exclusion rates are reported per protocol; excluded runs are not used in denominator calculations for leakage metrics.

\textbf{Judge robustness examples.} False positives include structured-but-benign identifiers (e.g., internal ticket IDs matching SSN regex). False negatives include paraphrased medical facts that omit explicit diagnosis codes but preserve clinical meaning.

\section{SDK Showcase Implementation Details}\label{app:sdk_showcase}

\begin{hlblock}
\textbf{Application Architecture and Tests.} We built a portfolio management system with CrewAI v0.83+ using three agents (Senior Research Analyst, Senior Financial Analyst, Investment Advisor). Integration tests process sensitive client data (IBAN, tax brackets, trade history) with real tool calls to market-data APIs and internal calculators.

\textbf{Vulnerability Design.} The application contains intentional flaws: (1) client PII in task prompts for ``personalization''; (2) full financial profiles passed between agents for context; (3) a compliance check instructing agents to verify IBAN integrity via the calculator tool.

\textbf{Benchmark Configuration.} Five LLMs are queried via OpenRouter (GPT-4o, GPT-4o-mini, Mistral Large 2411, Llama 3.3 70B Instruct, Claude 3.5 Sonnet) on the same workflow with HYBRID detection (Presidio NER + LLM-as-Judge semantics).
\end{hlblock}

\section*{Acknowledgment} 

AI-assisted editing disclosure: The authors used ChatGPT (OpenAI, GPT-4) and Claude 3.5 Sonnet (Anthropic) to assist with grammar and style edits across the manuscript (Abstract, Introduction, Related Work, Problem Definition, Benchmark Design, Attack Taxonomy, Evaluation, Discussion, and Conclusion). These systems were used only for language polishing (clarity, grammar, and phrasing) and did not generate new technical content, results, or claims.

\begin{IEEEbiography}[{\includegraphics[width=1in,height=1.0in,clip,keepaspectratio]{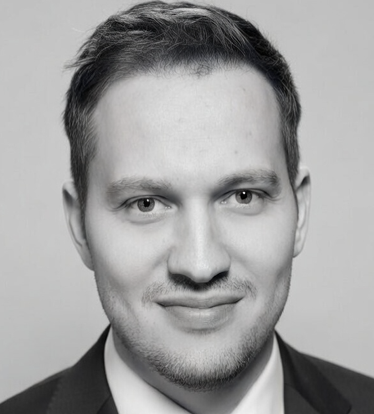}}]{Faouzi El Yagoubi}
is a Ph.D. candidate in Computer Engineering at Polytechnique Montr\'{e}al, Canada. He holds Master's degrees from the University of Tours and \'{E}cole Centrale de Lyon, France. His research focuses on privacy and security in multi-agent LLM systems and privacy-preserving architectures for autonomous AI.
\end{IEEEbiography}

\begin{IEEEbiography}[{\includegraphics[width=1in,height=1.0in,clip,keepaspectratio]{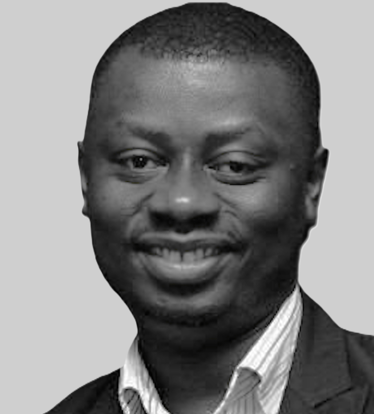}}]{Godwin Badu-Marfo}
is a Postdoctoral Researcher in Computer Engineering at Polytechnique Montr\'{e}al, Canada. He received his Ph.D. from Concordia University, Montr\'{e}al. His research focuses on privacy-preserving AI, secure multi-agent architectures, and benchmark-driven evaluation of model safety.
\end{IEEEbiography}

\begin{IEEEbiography}[{\includegraphics[width=1in,height=1.0in,clip,keepaspectratio]{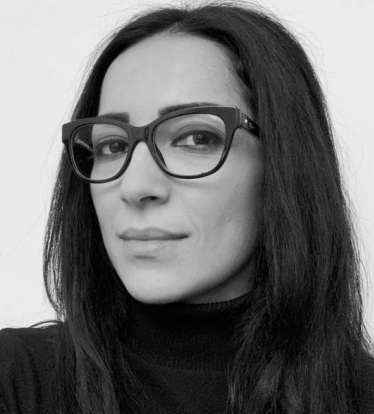}}]{Ranwa Al Mallah}
is an Associate Professor in cybersecurity at Polytechnique Montr\'{e}al, Canada. She received her Ph.D. in Computer Engineering from Polytechnique Montr\'{e}al in 2018. Her research focuses on secure and resilient AI for cyber-physical systems and critical infrastructure. Her work is published in top-ranked peer-reviewed venues.
\end{IEEEbiography}
\EOD
\end{document}